\documentclass{article}
\usepackage{iclr2027_conference,times}

\usepackage{amsmath,amsfonts,bm}

\def\eqref#1{equation~\ref{#1}}

\def\1{\bm{1}}

\DeclareMathAlphabet{\mathsfit}{\encodingdefault}{\sfdefault}{m}{sl}
\SetMathAlphabet{\mathsfit}{bold}{\encodingdefault}{\sfdefault}{bx}{n}

\usepackage{hyperref}
\usepackage{url}
\usepackage{booktabs}
\usepackage{threeparttable}
\usepackage{multirow}
\usepackage{tabularx}
\usepackage{array}
\usepackage{caption}
\usepackage[table]{xcolor}
\usepackage{graphicx}
\definecolor{oursgreen}{RGB}{234,247,234}
\definecolor{stamrow}{RGB}{237,246,239}
\usepackage[detect-weight,mode=text]{siunitx}
\usepackage{float}

\definecolor{stamrow}{RGB}{237,246,239}

\newcommand{\metrichead}[2]{%
  \shortstack{#1\\[-1pt]\scriptsize #2}%
}

\title{Personalized State-Transition-Aware Memory for Clinical Agents}

\author{
Maryam Haghifam$^{1}$ \quad
Zahra Rajabi$^{2}$ \quad
Yizhou Sun$^{1}$ \quad
Carlos Morato$^{2}$ \\
$^{1}$Department of Computer Science, University of California, Los Angeles, CA, USA \\
$^{2}$Optum AI, UnitedHealth Group, Minneapolis, MN, USA
}

\iclrfinalcopy

\begin{document}

\maketitle
\fancyhead{}

\begin{abstract}
Large language model (LLM) agents that reason over clinical records must track changes in a patient’s state while preserving the history needed to understand them. Simply accumulating memories leaves unclear which information still applies, whereas overwriting earlier memories can erase evidence needed to reconstruct treatment history and clinical trajectories. We introduce \textbf{STAM}, a state-transition-aware memory framework that records state changes as new clinical entries arrive. STAM combines semantic retrieval with typed clinical relations to identify affected memories, maintaining current information in \textsc{Active} and superseded or resolved information in \textsc{History}. At read time, a query-dependent gate selectively serves historical memory. Across four longitudinal clinical benchmarks, we evaluate STAM with downstream
question answering, direct state-maintenance diagnostics, and comparisons at matched
evidence lengths. On MedMemoryBench, the full write-time pipeline increases supersession-pair recall from 28.6 to 53.9 and reduces false archival relative to deterministic state-maintenance rules. At matched evidence lengths, STAM also shows six significant improvements and no significant decreases across paired comparisons with memory and retrieval baselines. These findings support evaluating longitudinal memory systems for how they manage state as well as how they retrieve information.
\end{abstract}

\section{Introduction}
\label{sec:intro}

Long-term memory enables large language model (LLM) agents to retain and use information across interactions~\citep{xu2025amem,chhikara2025mem0,li2025memos}. A patient's record grows over visits and admissions: medications change, conditions resolve, and later measurements may make earlier values outdated for questions about the current state~\citep{cui2025timer,wang2026medmemorybench}. A clinical memory system must identify which information still describes the patient's current state while preserving earlier information needed to understand treatment history and clinical trajectories. As records grow, providing the full history to the answering model may become costly or exceed its context limit. The system must therefore preserve useful history while selecting the evidence needed for each question.

\begin{figure*}
    \centering
    \includegraphics[width=0.9\textwidth]{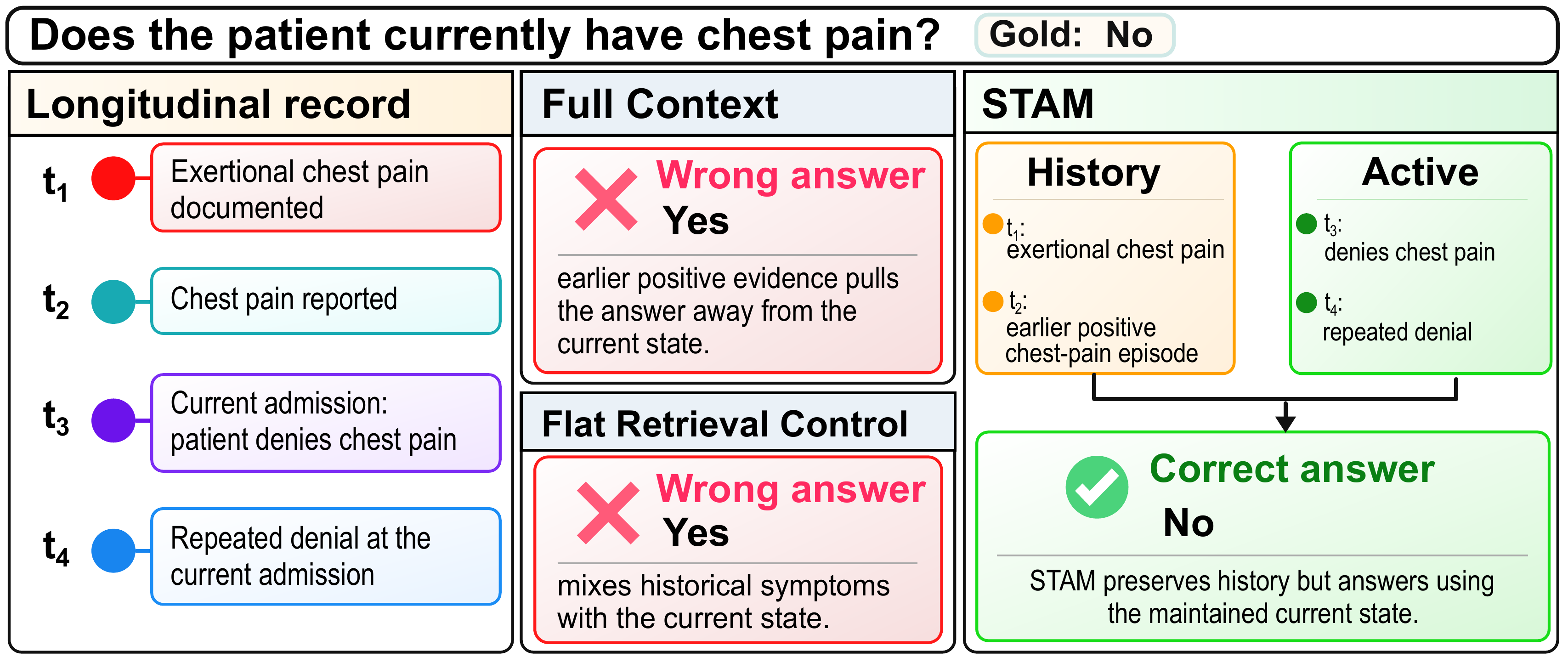}
\caption{Illustrative example of write-time state maintenance. In this example,
Full Context and flat retrieval mix earlier positive chest-pain evidence with
the current state and answer incorrectly.}
    \label{fig:stam_intro}
\end{figure*}

Semantic similarity alone does not establish whether a memory still describes the patient's current state. If a patient's insulin dose is increased, for example, similarity-based retrieval may return both the earlier and updated doses in response to a question about the current regimen. The earlier dose remains part of the treatment history, but using it to answer a question about the current regimen gives an incorrect answer. Consistent with this concern, MedCache reports outdated evidence as the largest error category for seven of the eight systems in its error analysis~\citep{chan2026medcache}. It also documents a case in which three systems return an earlier laboratory measurement from the same admission when asked for the most recent value. Figure~\ref{fig:stam_intro} illustrates this failure mode and how explicit
state maintenance separates earlier evidence from the current state.

Historical information remains necessary for questions about treatment changes,
earlier admissions, and clinical trajectories. However, temporal order alone
does not establish whether a new observation supersedes, qualifies, or coexists
with an earlier fact. The time at which information is recorded may also differ
from the period to which it applies~\citep{snodgrass1985taxonomy}. Even when
event times are known, clinical content matters: a new medication may supplement
an existing regimen rather than replace it, and a treatment change may affect
whether a related monitoring plan still applies. Determining which earlier
memories are affected therefore requires considering both their content and
their relationships to new evidence. These cases motivate evaluating write-time
state maintenance directly, alongside downstream question answering, when only
part of the longitudinal record can be provided to the model.

We introduce \textbf{STAM}, a state-transition-aware memory framework for
longitudinal clinical agents that records how patient information changes over
time. STAM combines semantic retrieval with graph-based discovery over typed
clinical relations to identify memories that may be affected by new evidence.
At each update, currently applicable memories remain in \textsc{Active}, while
superseded or resolved information is preserved in \textsc{History} with a
reason for the change and, when applicable, a link to the newer memory. When answering, STAM
uses the question to decide whether historical memories should be included in
the evidence. Permanent deletion is handled separately by a conservative
deletion gate. 

Across four longitudinal clinical benchmarks, we evaluate STAM through both
direct state-maintenance diagnostics and downstream question answering.
Write-time diagnostics show improved supersession tracking and lower false
archival relative to simpler state-maintenance controls. Under a common reader
protocol and approximately matched context lengths, STAM shows significant
gains in several comparisons with memory and retrieval baselines, with no
significant decreases.

\section{Related Work}

\subsection{Long-Term Memory Systems for LLM Agents}

Long-term memory systems extend LLM agents beyond a fixed interaction context by storing and retrieving information from prior interactions. A-MEM organizes memories through indexing, linking, and memory evolution~\citep{xu2025amem}, while Mem0 extracts and consolidates salient information and supports graph-based conflict handling~\citep{chhikara2025mem0}. LightMem targets efficient memory construction through hierarchical memory and sleep-time updates~\citep{fang2026lightmem}, and MemOS manages multiple memory types as system resources~\citep{li2025memos}. MemRL learns episodic-memory utility through reinforcement learning~\citep{zhang2026memrl}, while Memory-R1 learns ADD, UPDATE, DELETE, and NOOP operations~\citep{yan2026memoryr1}. These systems primarily address general memory organization, retrieval, and
updating. STAM focuses on maintaining changing clinical state: it identifies
earlier memories that may be affected by new evidence while preserving prior
states needed for historical questions.

Our read-time mechanisms also relate to adaptive retrieval and forgetting.
Self-RAG retrieves evidence on demand, while Adaptive-RAG selects retrieval
strategies based on query complexity~\citep{asai2024selfrag,jeong2024adaptiverag}.
Our store gate instead decides whether historical memory should be included in
the evidence for a query.

\subsection{Temporal Validity and Evolving State in Agent Memory}

Several systems represent changing information explicitly. Zep uses a temporal
knowledge graph for agent memory~\citep{rasmussen2025zep}, while TSM models
semantic time and durative memory~\citep{su2026tsm}. APEX-MEM preserves
information in append-only storage and resolves changes at query
time~\citep{banerjee2026apexmem}. MemStrata uses deterministic supersession in
a bitemporal ledger~\citep{yadav2026memstrata}, while TOKI uses bitemporal
operators to resolve contradictions and retains overridden facts in audit
records~\citep{wang2026toki}. Other work directly studies whether stored information remains valid as state
changes. STALE examines implicit conflicts and introduces CUPMem, which performs
write-time state revision with propagation-aware candidate search
~\citep{chao2026stale}. StateMem tracks supersession and relational
dependencies between state units~\citep{fan2026statemem}. A-TMA assigns
current, historical, and transition roles to memories, constructs evidence
packets for the state requested by a query, and evaluates failures separately
at memory maintenance, retrieval, and answering~\citep{shi2026atma}. Memora
introduces an evaluation metric that penalizes reliance on obsolete or
invalidated memories~\citep{uddin2026memora}. These works establish evolving-state tracking as an important direction in
agent memory. STAM targets longitudinal clinical records, using
patient-specific typed clinical relations together with semantic retrieval
to identify affected memories and directly evaluating the resulting
write-time state decisions.

\subsection{Longitudinal Clinical Memory and Reasoning}

Longitudinal clinical reasoning has been studied through temporal modeling
and memory-oriented benchmarks
\citep{cui2025timer,xu2026longmedbench,stinard2026clinicalbench,
wang2026medmemorybench,zhang2026medlocomo,wang2026clintracebench}.

MedCache is closely related to our setting. It represents clinical facts with
validity intervals, preserves bounded trajectories for time-sensitive concepts,
organizes evidence into overlapping specialty memories, and reconstructs the
applicable patient state at query time~\citep{chan2026medcache}. Its design
emphasizes temporal curation, specialty-based memory organization, and query
routing. STAM instead focuses on write-time identification of clinically
related memories whose current status may change: semantic retrieval and typed
clinical relations jointly identify affected memories, after which the system
records the resulting state changes. We evaluate this write-time process
directly, alongside downstream question answering under controlled evidence
lengths.
\section{Method}
\label{sec:method}
\begin{figure}[t]
    \centering
    \includegraphics[width=0.9\linewidth]{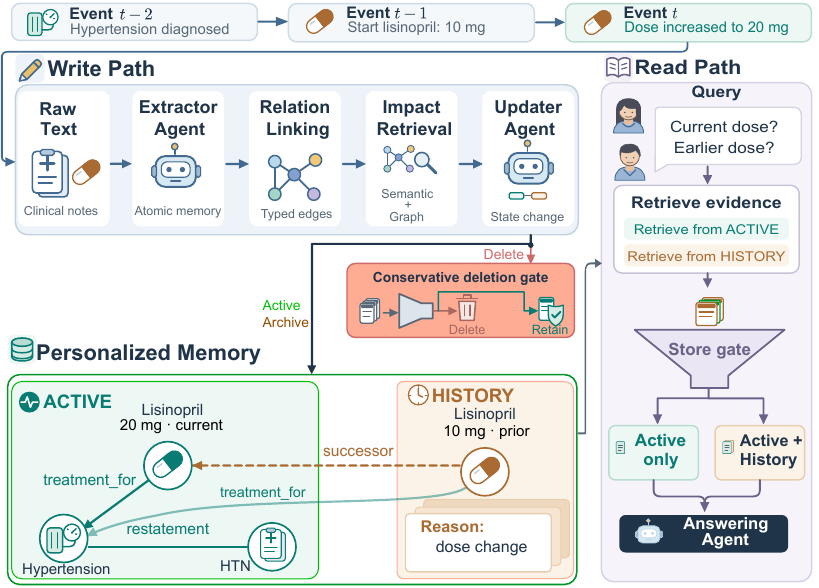}
    \caption{Overview of the proposed state-aware memory framework.}
    \label{fig:framework}
\end{figure}

\subsection{Setting and Framework Overview}
\label{sec:method_overview}

Let $t$ index record entries in the order they become available to the system.
After processing $t$ entries, STAM represents the personalized memory state as
$\mathcal{M}_t=(\mathcal{A}_t,\mathcal{H}_t,\mathcal{G}_t)$, where
$\mathcal{A}_t$ and $\mathcal{H}_t$ denote the \textsc{Active} and
\textsc{History} stores, respectively, and $\mathcal{G}_t$ denotes the
relation graph. The index $t$ denotes update order and need not coincide with
the clinical time described by an entry. Given a newly available record entry
$e_t$, the write path updates $\mathcal{M}_{t-1}$ to $\mathcal{M}_t$.

The personalized memory is organized into two stores: \textsc{Active}
represents the maintained current clinical state, while \textsc{History}
preserves superseded or resolved memories.

\subsection{Write Path: Memory Construction and State Update}
\label{sec:write_path}

The write path processes record entries as they become available. A record
entry may contain free-text notes, laboratory measurements, medication
changes, orders, or other encounter information. Processing proceeds through
four stages: atomic memory extraction, relation linking, impact retrieval, and
state update. Rather than reconsidering the entire memory after every new
entry, STAM identifies prior memories that may be affected and restricts state
reasoning to this candidate set.

\paragraph{Atomic memory extraction.}
Given an incoming record entry $e_t$, the extractor agent converts it into a set of atomic memories $\mathcal{E}_t=\{m_t^1,\ldots,m_t^{N_t}\}$, where $\mathcal{E}_t$ denotes the memories extracted from event $e_t$, $N_t$ is their number, and $m_t^i$ denotes the $i$-th atomic memory. Each atomic memory represents a single person-specific clinical assertion rather than an entire note or encounter.

\paragraph{Relation linking.}
After atomic memory extraction, the system integrates the newly extracted memories into the person-specific relation graph. Each new memory is linked to relevant \textsc{Active} memories using typed edges. We use four undirected relation types, \texttt{restatement}, \texttt{same\_condition\_thread}, \texttt{drug\_interaction}, and \texttt{systemic\_link}, and three directed relation types, \texttt{causal}, \texttt{treatment\_for}, and \texttt{monitoring\_for}. 

The resulting graph provides a complementary discovery channel during impact retrieval. When a memory transitions to \textsc{History}, its existing graph relations are preserved, but no new relations are subsequently added to that historical memory. These relation types define the clinical configuration used in our experiments;
the framework itself does not require this fixed relation vocabulary.

\paragraph{Impact retrieval.}
Given the newly extracted memories $\mathcal{E}_t$, we identify prior memories whose temporal validity may be affected by the new clinical information. Impact retrieval differs from query-time retrieval: rather than retrieving evidence for answering a question, it identifies existing memories whose temporal status may need to be updated. The search considers both the \textsc{Active} and \textsc{History} stores, allowing affected memories to be identified across both current and prior states.

We combine semantic retrieval with graph-based discovery. Semantic retrieval identifies prior memories related to $\mathcal{E}_t$ in embedding space. For graph-based discovery, we expand one hop from each newly linked memory and score neighboring memories by their strongest relation to any memory in $\mathcal{E}_t$. The two channels serve complementary roles: semantic retrieval primarily surfaces directly affected memories, which are often close to the arriving evidence in embedding space, whereas graph-based discovery can capture dependency-mediated effects that are less semantically similar. For example, discontinuing a treatment may require reconsidering related memories about its monitoring plan or interaction constraints even when those memories share little surface language with the new event. We take the union of candidates returned by the two channels and denote the resulting impact candidate set by $\mathcal{C}^{\mathrm{imp}}_t$.

\paragraph{Memory state update.}
The update agent is a language model invoked once per decision with a fixed prompt and no tools. For each new atomic memory and its candidates in $C_t^{\mathrm{imp}}$,
the agent decides whether to incorporate it, whether it describes a
current or prior state, and whether affected \textsc{Active} memories
remain current. Decisions use clinical content and temporal relations rather than
ingestion order alone. If its content refers to an earlier clinical state, it may be incorporated directly into \textsc{History}. Conversely, an existing \textsc{Active} memory remains active unless the incoming evidence establishes that it has been superseded, resolved, or otherwise no longer describes the current state.

The central mechanism is temporal state transition. When new evidence indicates that an existing \textsc{Active} memory no longer describes the current state, that memory is moved to \textsc{History} rather than deleted or left to compete with newer evidence as current information. For each transition, STAM records why the memory's status changed and links it to the newer memory when one exists. The updater also optionally emits a \textsc{Delete} proposal for a memory
that no longer appears useful for future reasoning. This proposal does not
itself remove the memory: without the deletion gate, such memories are retained
in \textsc{History}. Permanent removal is decided separately by the conservative
deletion gate in Section~\ref{sec:deletion}. The two stores are not isolated: preserved relations and successor links allow a clinical thread to span \textsc{Active} and \textsc{History}.

\subsection{Read Path: State-Aware Evidence Selection}
\label{sec:read_path}

Store assignment records the updater's judgment about whether a memory
describes the current state. Given a question $q$, the retriever searches
\textsc{Active} and \textsc{History} separately under their respective
retrieval budgets, producing
$\mathcal{C}^{A}_q=\mathcal{R}(q,\mathcal{A}_t)$ and
$\mathcal{C}^{H}_q=\mathcal{R}(q,\mathcal{H}_t)$.
The store gate then determines whether the answering model receives only
\textsc{Active} candidates or candidates from both stores.
Appendix~\ref{app:store_aware_retrieval_controls} evaluates joint ranking and
the independent effect of exposing store labels.



\subsection{Selective History Serving}

Some questions require prior-state evidence, whereas others can be answered
from \textsc{Active} alone. The store gate uses query--memory similarity
signals to choose between \textsc{Active}-only and
\textsc{Active}+\textsc{History} serving. The benchmarks do not annotate
whether a question requires \textsc{History}, so agreement between the two
serving arms is used as a proxy for this decision. The policy therefore does
not require labels for History necessity; its threshold is selected using
replayed answer quality on the training folds and is fixed before evaluation.
We evaluate multiple values of $\lambda$, which controls the trade-off between
answer quality and serving efficiency
(Appendix~\ref{app:store_gate}).

\subsection{Conservative Memory Deletion}
\label{sec:deletion}

Because \textsc{History} grows over longitudinal follow-up, retaining every
historical memory increases storage and retrieval space. Since a superseded
memory may still be useful, STAM separates state updating from permanent
removal and applies a conservative deletion gate only to archived memories.
Label construction, calibration, and deletion diagnostics are reported in
Appendix~\ref{app:deletion_gate}.
\section{Experimental Setup}
\label{sec:experiment_setup}

\paragraph{Datasets.}
We evaluate on four clinical benchmarks spanning synthetic longitudinal interactions,
MIMIC-grounded synthetic medical dialogues, structured EHR event sequences,
and clinical notes. MedMemoryBench~\citep{wang2026medmemorybench} contains synthetic longitudinal healthcare interactions in which person-specific information accumulates across sessions; we evaluate 496 questions spanning six query types, including temporal localization, state update, and multi-hop clinical deduction. MedLoCoMo~\citep{zhang2026medlocomo} contains synthetic doctor--patient
dialogues grounded in MIMIC-IV and MIMIC-IV-Note records; we evaluate a
320-question subset including cross-admission comparison, longitudinal
progression, and adversarial questions. MIMIC-QA (ours) is constructed deterministically from
MIMIC-IV~\citep{johnson2023mimiciv}. It contains 2,000 questions over
100 patients, with a patient-disjoint 1,400/200/400 train/validation/test split. ClinicalBench~\citep{stinard2026clinicalbench} contains 400 questions over 43 patients grounded in MIMIC-IV clinical notes and spans nine assertion-sensitive question categories. Across the evaluated cohorts, mean per-person record length varies substantially: 719K tokens on MedMemoryBench, 37.9K on MedLoCoMo, 64.7K on MIMIC-QA, and 20.0K on ClinicalBench. Additional dataset statistics, sampling procedures, context-length
measurements, and construction details are provided in Appendix~\ref{app:implementation}.

\paragraph{Models and comparison setup.}
Our local write path uses Qwen3-1.7B~\citep{yang2025qwen3}, final answer
generation uses Qwen3-4B-Instruct-2507~\citep{yang2025qwen3}, and dense
retrieval uses BGE-small-en-v1.5~\citep{xiao2023cpack}. We report the wall-clock and model-compute cost of constructing and maintaining STAM's memory in Appendix~\ref{app:write-cost}. GPT-5 calls use the fixed \texttt{gpt-5-2025-08-07} snapshot;
memory-construction backbones and baseline-specific adaptations
are detailed in Appendix~\ref{app:implementation}. Full Context serves the temporally ordered raw longitudinal record up to the
reader-specific context limit, truncating the oldest record entries first when
necessary. Thus, ``Full Context'' denotes the maximum admissible raw context rather than necessarily the complete record. On MedMemoryBench, 329 of the 496 GPT-5 Full Context inputs require
truncation. MIMIC-QA also requires truncation for a subset of questions,
whereas MedLoCoMo and ClinicalBench fit without truncation. Full truncation
statistics and the MedMemoryBench context-volume analysis are provided in
Appendix~\ref{app:full_context_truncation}. Flat RAG retrieves semantically ranked chunks without structured memory construction. Baseline-specific implementation
details are provided in Appendix~\ref{app:implementation}.


\paragraph{Evaluation metrics.}
We use benchmark-specific scoring protocols. For MedMemoryBench, we report
macro accuracy, using deterministic scoring for entity exact match and
multiple-choice questions and GPT-5 judging for the remaining question types.
For MedLoCoMo, we report type-macro F1, obtained by averaging per-question
scores within each of the six question types and then averaging across types.
For MIMIC-QA, we report mean token-level F1 over questions. For ClinicalBench,
we report pooled accuracy using our GPT-5 judge against the reference answers
used in our evaluation. Question-type-specific scoring details are provided in
Appendix~\ref{app:dataset_protocols}.
\begin{table}[t]
\centering
\caption{
Answer quality (\%; higher is better) across persistent memory systems
with Qwen3-4B-Instruct and GPT-5 readers.
}
\label{tab:main_results}

\small
\setlength{\tabcolsep}{3pt}
\renewcommand{\arraystretch}{1.05}
\medskip

\begin{minipage}[t]{0.49\linewidth}
\vspace{0pt}
\centering
\textbf{(a) Qwen3-4B-Instruct}\par\smallskip

\begin{tabularx}{\linewidth}{l*{4}{>{\centering\arraybackslash}X}}
\toprule
Method
& \metrichead{MMB}{mAcc.}
& \metrichead{MLC}{mF1}
& \metrichead{MQA}{F1}
& \metrichead{CB}{Acc.} \\
\midrule
Graphiti (Zep) & 26.8 & 31.9 & 23.3 & 47.5 \\
LightMem       & 35.7 & 33.2 & 22.2 & 50.5 \\
A-Mem          & 32.5 & 39.0 & 42.3 & 49.2 \\
MemOS          & 32.0 & 27.5 & 19.9 & 46.8 \\
Mem0           & 22.1 & 28.3 & 23.1 & 45.2 \\
MemRL          & 21.6 & 22.7 & 22.2 & 47.0 \\
CUPMem         & 5.7  & 19.3 & 7.1  & 47.2 \\
\midrule
\rowcolor{stamrow}
\textbf{STAM}
& 37.4 & 36.2 & 57.4 & 50.2 \\

\addlinespace[2pt]
\rowcolor{stamrow}
\textbf{STAM + FT}
& 45.5 & 52.1 & 90.2 & 70.7 \\
\rowcolor{stamrow}
\textbf{STAM + Write FT}
& 39.3 & 39.8 & 56.2 & 53.8 \\
\bottomrule
\end{tabularx}
\end{minipage}
\hfill
\begin{minipage}[t]{0.49\linewidth}
\vspace{0pt}
\centering
\textbf{(b) GPT-5}\par\smallskip

\begin{tabularx}{\linewidth}{l*{4}{>{\centering\arraybackslash}X}}
\toprule
Method
& \metrichead{MMB}{mAcc.}
& \metrichead{MLC}{mF1}
& \metrichead{MQA}{F1}
& \metrichead{CB}{Acc.} \\
\midrule
Graphiti (Zep) & 49.2 & 35.6 & 28.9 & 43.0 \\
LightMem       & 49.4 & 30.9 & 25.6 & 48.8 \\
A-Mem          & 48.9 & 37.9 & 43.7 & 52.2 \\
MemOS          & 46.9 & 34.6 & 10.9 & 50.0 \\
Mem0           & 37.1 & 29.9 & 19.2 & 51.8 \\
MemRL          & 28.1 & 21.9 & 9.5  & 44.8 \\
CUPMem         & 20.4 & 17.3 & 2.9  & 35.0 \\
\midrule
\rowcolor{stamrow}
\textbf{STAM}
& 52.9 & 36.9 & 87.2 & 49.0 \\
\bottomrule
\end{tabularx}
\end{minipage}

\end{table}

\section{Results}
\label{sec:results}
We evaluate write-time state maintenance, end-to-end question answering,
performance on state- and time-oriented questions, and the effects of
evidence serving and selective historical access.

\subsection{Comparison with Memory Systems}
\label{sec:main_results}

Table~\ref{tab:main_results} reports system-level answer quality under the
Qwen3-4B-Instruct and GPT-5 readers. Because memory systems can serve
different amounts of evidence to the reader, we additionally compare them
under the approximate context-matching protocol described in
Appendix~\ref{app:context-matched}.

Under the Qwen3-4B context-matching protocol, six of 17 paired comparisons
significantly favor STAM, with no significant decreases. These include a $+7.15$ Token-F1 difference
over Flat RAG on MIMIC-QA, significant gains over LightMem, MemOS, and MemRL
on MIMIC-QA, and gains over MemRL on MedMemoryBench and MedLoCoMo.
Full per-system results and significance tests are reported in
Appendix~\ref{app:context-matched}.

STAM + Write FT fine-tunes only the write modules using benchmark-specific
LoRA adapters, whereas STAM + FT additionally fine-tunes the reader
(Appendix~\ref{app:write_finetuning}). Full Context and Flat RAG do not
maintain an evolving structured memory and are therefore analyzed separately
as reference strategies in Section~\ref{sec:reference_strategies}.

\subsection{State- and Temporal-Question Analysis}
\label{sec:state_temporal_results}

We focus on question categories that directly test state updating,
longitudinal progression, current-state identification, or event sequence,
which are the operations most directly aligned with STAM's state-maintenance
objective.

\begin{table*}[t]
\centering
\small

\begin{minipage}[t]{0.46\linewidth}
\centering
\textit{(a) State/time categories}\\[2pt]

\setlength{\tabcolsep}{2.4pt}
\renewcommand{\arraystretch}{1.02}

\begin{tabular}{@{}llrrrr@{}}
\toprule
& & \multicolumn{2}{c}{Qwen}
  & \multicolumn{2}{c}{GPT-5} \\
\cmidrule(lr){3-4}
\cmidrule(lr){5-6}

Data & Type
& STAM & FC
& STAM & FC \\
\midrule

MMB
& State update
& \cellcolor{stamrow}42.0 & 34.0
& \cellcolor{stamrow}68.0 & 66.0 \\

MLC
& Longitudinal
& \cellcolor{stamrow}21.9 & 14.8
& \cellcolor{stamrow}24.8 & 28.9 \\

CB
& Current state
& \cellcolor{stamrow}56.0 & 50.0
& \cellcolor{stamrow}66.0 & 62.0 \\

&
Sequence
& \cellcolor{stamrow}47.5 & 20.0
& \cellcolor{stamrow}50.0 & 45.0 \\

\bottomrule
\end{tabular}
\end{minipage}
\hfill
\begin{minipage}[t]{0.52\linewidth}
\centering
\textit{(b) Overall QA}\\[2pt]

\setlength{\tabcolsep}{2.3pt}
\renewcommand{\arraystretch}{1.02}

\begin{tabular}{@{}lccc ccc@{}}
\toprule
& \multicolumn{3}{c}{Qwen}
& \multicolumn{3}{c}{GPT-5} \\
\cmidrule(lr){2-4}
\cmidrule(lr){5-7}

Data
& FC & RAG & STAM
& FC & RAG & STAM \\
\midrule

MMB
& 30.5 & 39.1 & \cellcolor{stamrow}37.4
& 63.2 & 55.3 & \cellcolor{stamrow}52.9 \\

MLC
& 37.4 & 32.2 & \cellcolor{stamrow}35.7
& 42.5 & 32.6 & \cellcolor{stamrow}36.9 \\

MQA
& 43.8 & 45.4 & \cellcolor{stamrow}52.6
& 85.6 & 56.7 & \cellcolor{stamrow}87.2 \\

CB
& 51.0 & 47.0 & \cellcolor{stamrow}50.0
& 51.8 & 47.0 & \cellcolor{stamrow}49.0 \\

\bottomrule
\end{tabular}
\end{minipage}

\caption{
(a) State- and time-oriented question categories.
(b) Overall comparison with Full Context (FC) and Flat RAG.
For the Qwen3-4B-Instruct comparisons with Flat RAG, STAM uses the
approximately context-matched configurations described in Appendix~\ref{app:context-matched}.
}
\label{tab:reference_and_state}
\end{table*}

Table~\ref{tab:reference_and_state}(a) compares STAM with raw Full Context
on state- and time-oriented question categories. With Qwen3-4B-Instruct,
STAM scores higher than Full Context on all four reported categories:
MedMemoryBench state update, MedLoCoMo longitudinal progression, and
ClinicalBench current-state and sequence questions.

With GPT-5, STAM scores higher on MedMemoryBench state-update questions
(68.0 vs.\ 66.0) and both ClinicalBench categories, but lower on
MedLoCoMo longitudinal progression (24.8 vs.\ 28.9).
The largest difference is on ClinicalBench sequence questions with
Qwen3-4B-Instruct (47.5 vs.\ 20.0, $p=.003$). The ClinicalBench
current-state difference is not statistically significant; the remaining
category comparisons are treated descriptively. Across these categories, STAM's relative performance is generally stronger
than its overall comparison with Full Context, suggesting that the aggregate
Full Context advantage is not concentrated in the state- and time-oriented
questions examined here.

\subsection{Evidence Serving and Selective History}
\label{sec:serving_results}
\paragraph{Serving-budget analysis.}
With GPT-5, serving depth and evidence selection substantially affect
downstream performance. On MedMemoryBench, relaxing the bounded serving
restriction raises macro accuracy from 52.90 under the default configuration
to 63.84 when all maintained memories are served, compared with 63.17 for
raw Full Context. This diagnostic comparison is not context matched, and
Full Context is truncated for 329 of 496 questions.

On MedLoCoMo, where the complete raw record fits in context, a 4,500-word
ranked condition reaches 40.75 type-macro F1, compared with 42.53 for
Full Context while serving 4,494 versus 25,157 median words. The ranked
condition also outperforms serving all maintained memories in record order
(38.80), showing that evidence selection matters in addition to serving
volume. Full sweeps are reported in
Appendix~\ref{app:gpt5-serving-budget}.

\paragraph{Selective History serving.}
The Store Gate routes selected queries to ACTIVE-only serving. Across the selected local-reader configurations, the Store Gate omits
HISTORY for 1.0\%--79.0\% of queries, with no statistically significant
change in answer quality. Omitting HISTORY also reduces prompt length and
prefill TTFT at the serving-arm level.
Full routing, quality, and latency results are reported in
Appendices~\ref{app:store_gate} and
\ref{app:serving_efficiency}.

\subsection{Reference Strategies and Memory Controls}
\label{sec:reference_strategies}
Panel (b) of Table~\ref{tab:reference_and_state} compares STAM with raw
Full Context and Flat RAG. For the Qwen3-4B-Instruct comparisons with
Flat RAG, STAM uses the approximately context-matched configurations
described in Appendix~\ref{app:context-matched}. STAM scores higher than
Full Context on MedMemoryBench (37.4 vs.\ 30.5) and MIMIC-QA
(52.6 vs.\ 43.8), but lower on MedLoCoMo (35.7 vs.\ 37.4) and
ClinicalBench (50.0 vs.\ 51.0). Compared with Flat RAG, STAM is lower
on MedMemoryBench (37.4 vs.\ 39.1) and higher on the other three
benchmarks.

With GPT-5, Full Context scores higher than STAM on MedMemoryBench,
MedLoCoMo, and ClinicalBench. STAM scores higher on MIMIC-QA
(87.2 vs.\ 85.6 Token-F1), although this difference is not statistically
significant ($p=.18$). Overall, the relationship between memory
and raw Full Context varies by dataset and reader rather than showing a
uniform advantage in either direction.

\paragraph{Conservative deletion.}
At the calibrated threshold, the deletion gate removes 2.3\% of stored
memories on MedMemoryBench with 92.9\% safe-to-forget precision and 0.3\%
never-forget loss, measured against LLM-derived endurance labels. In paired
downstream evaluation, 20 answers improve and 14 regress ($p=.39$),
providing no evidence of aggregate QA degradation. Full calibration and
retention analyses are reported in Appendix~\ref{app:deletion_gate}.
\section{Ablation Studies}
\label{sec:ablations}

\subsection{Historical Memory Access}
\label{sec:store_retrieval_ablations}
\label{sec:abl_history}

\begin{table*}[t]
\centering
\small
\setlength{\tabcolsep}{6pt}
\renewcommand{\arraystretch}{1.08}

\begin{tabular}{lrrr@{\hspace{10pt}}rrr}
\toprule
& \multicolumn{3}{c}{Qwen3-4B-Instruct}
& \multicolumn{3}{c}{GPT-5} \\
\cmidrule(lr){2-4}\cmidrule(lr){5-7}
Dataset
& A+H & ACTIVE only & $\Delta$
& A+H & ACTIVE only & $\Delta$ \\
\midrule
MMB
& 35.68 & 34.12 & +1.57
& 51.27 & 46.27 & +5.00 \\

MLC
& 35.91 & 36.00 & $-0.09$
& 36.73 & 38.28 & $-1.56$ \\

MQA
& 54.73 & 52.63 & +2.10
& 85.26 & 53.54 & +31.72 \\

CB
& 50.50 & 49.75 & +0.75
& 49.50 & 50.00 & $-0.50$ \\
\bottomrule
\end{tabular}

\caption{
Historical-access ablation. Comparisons are paired within each
dataset/model configuration; A+H serves both stores, whereas ACTIVE only
omits \textsc{History}.
}
\label{tab:history_crossdataset}
\end{table*}

\paragraph{Does access to \textsc{History} improve answer quality?}
The contribution of historical evidence depends on the dataset and answering
configuration (Table~\ref{tab:history_crossdataset}). With Qwen3-4B-Instruct,
serving \textsc{Active}+\textsc{History} improves performance on
MedMemoryBench, MIMIC-QA, and ClinicalBench, while the difference on
MedLoCoMo is negligible. With GPT-5, historical access improves
MedMemoryBench and MIMIC-QA, but slightly reduces performance on
MedLoCoMo and ClinicalBench. Thus, historical evidence can be useful, but
its contribution is not uniform across settings.

To separate historical content from a reduction in serving volume, we
additionally use a matched-volume refill control that replaces removed
\textsc{History} with additional \textsc{Active} memories
(Appendix~\ref{app:history_requirement}). Full refill, significance, and
routing analyses are reported in Appendix~\ref{app:store_gate}.

\subsection{Write-Time State Maintenance}
\label{sec:abl_state_maintenance}

\paragraph{Does state-aware updating help beyond append-only storage?}
We compare STAM with an append-only variant that retains historical
information but does not perform state transitions
(Table~\ref{tab:append_only}). STAM is numerically higher on all four
benchmarks. Because both configurations retain earlier information, this
comparison tests whether state-aware updating provides benefit beyond
simply accumulating memories.

\begin{table}[t]
\centering
\small
\setlength{\tabcolsep}{6pt}
\renewcommand{\arraystretch}{1.05}

\begin{tabular}{lcccc}
\toprule
Method
& MMB
& MLC
& MQA
& CB \\
\midrule

Append-only
& 34.16
& 35.40
& 55.99
& 49.25 \\

\rowcolor{stamrow}
\textbf{STAM}
& 37.40
& 36.20
& 57.45
& 50.20 \\

\midrule
$\Delta$
& +3.24
& +0.80
& +1.46
& +0.95 \\

\bottomrule
\end{tabular}

\caption{
Paired comparison with append-only memory under the Qwen3-4B-Instruct
reader using the append-only ablation configuration.
$\Delta$ denotes STAM minus append-only in score points.
MMB: macro accuracy; MLC: type-macro F1; MQA: Token-F1; CB: accuracy.
}
\label{tab:append_only}
\end{table}

To examine how the write path changes the maintained memory state, we
evaluate supersession-pair recall and false archival on MedMemoryBench,
and archive composition, evidence placement, and downstream QA on
MIMIC-QA.

\begin{table*}[t]
\centering
\small

\begin{minipage}[t]{0.48\linewidth}
\centering
\textit{(a) MedMemoryBench}\\[2pt]

\setlength{\tabcolsep}{3pt}
\renewcommand{\arraystretch}{1.0}
\begin{tabular}{lrr}
\toprule
Variant
& \shortstack{Pair\\recall $\uparrow$}
& \shortstack{False\\archival $\downarrow$} \\
\midrule
Full model
& 53.9
& 7.6 \\
Semantic only, $B=48$
& 51.6
& 12.3 \\
Rule: all non-episodic
& 28.6
& 17.6 \\
\bottomrule
\end{tabular}
\end{minipage}
\hfill
\begin{minipage}[t]{0.48\linewidth}
\centering
\textit{(b) MIMIC-QA}\\[2pt]

\setlength{\tabcolsep}{2.4pt}
\renewcommand{\arraystretch}{1.05}
\begin{tabular}{@{}lrrr@{}}
\toprule
Variant
& Archived (\%)
& \shortstack{Gold evidence\\in \textsc{History}}
& Token-F1 \\
\midrule
Full model
& 23.6
& 253
& 55.3 \\
\midrule
\shortstack[l]{No relations}
& 63.3
& 531
& 52.1 \\
\bottomrule
\end{tabular}
\end{minipage}

\caption{
Write-time ablations: (a) MedMemoryBench and (b) MIMIC-QA.
$B$ is the impact-candidate budget.
}
\label{tab:ablation_write}
\end{table*}

\paragraph{Why combine semantic and graph-based impact discovery?}
On MedMemoryBench, at the same impact-candidate budget, the full pipeline
achieves 53.9\% supersession-pair recall with 7.6\% false archival,
compared with 51.6\% recall and 12.3\% false archival using semantic
retrieval alone. The graph-based channel therefore provides additional
candidate information beyond semantic similarity, with the clearest
difference appearing in false archival.

\paragraph{Why not use a deterministic supersession rule?}
The deterministic rule that treats all non-episodic concepts as
supersedable reaches 28.6\% supersession-pair recall with 17.6\% false
archival, compared with 53.9\% recall and 7.6\% false archival for the
full write-time pipeline. The full pipeline therefore captures substantially
more annotated supersession relationships while making fewer false archival
decisions. Because this comparison changes the state-maintenance procedure
as a whole, we treat it as a pipeline-level control rather than attributing
the difference to a single component.

\paragraph{Why use typed clinical relations?}
On MIMIC-QA, removing typed relations increases the archived fraction from
23.6\% to 63.3\% and reduces Token-F1 from 55.3 to 52.1. The number of
reference evidence items placed in \textsc{History} also increases from
253 to 531. Because the no-relations variant archives substantially more
memories overall, this count should be interpreted together with the
archived fraction rather than by itself. These results show that typed
relations materially affect write-time state decisions and are associated
with higher downstream answer quality.

\section{Conclusion}

We presented STAM, a state-transition-aware memory framework that separates current from superseded or resolved information while preserving transitions between them. State-maintenance diagnostics show that semantic and relation-aware updating improves the organization of evolving state, while append-only comparisons show numerically higher downstream performance. At approximately matched evidence lengths, STAM shows several significant improvements and no significant decreases, while serving analyses show that performance also depends on which maintained memories are exposed to the reader. These results support evaluating longitudinal memory by both state maintenance and evidence serving.

\paragraph{Limitations.}
Our local write path is evaluated only with Qwen3-1.7B, so we do not characterize how state-maintenance quality changes with writer
capacity. The relation vocabulary is hand-specified, and deletion relies on
LLM-derived rather than clinician-validated endurance labels.

\bibliographystyle{iclr2027_conference}
\bibliography{iclr2027_conference}

@article{chao2026stale,
  title   = {{STALE}: Can {LLM} Agents Know When Their Memories Are No Longer Valid?},
  author  = {Chao, Hanxiang and Bai, Yihan and Sheng, Rui and Li, Tianle and Sun, Yushi},
  journal = {arXiv preprint arXiv:2605.06527},
  year    = {2026},
  url     = {https://arxiv.org/abs/2605.06527}
}

@article{shi2026atma,
  title   = {{A-TMA}: Decoupling State-Aware Memory Failures in Long-Term Agent Memory},
  author  = {Shi, Zitong and Tang, Yixuan and Tung, Anthony Kum Hoe},
  journal = {arXiv preprint arXiv:2607.01935},
  year    = {2026},
  url     = {https://arxiv.org/abs/2607.01935}
}

@article{fan2026statemem,
  title   = {Can Agent Memory Systems Track Evolving State?},
  author  = {Fan, Xinyi and Liu, Miri and Yang, Ruozhen and Ouyang, Siru and Han, Jiawei},
  journal = {arXiv preprint arXiv:2608.19652},
  year    = {2026},
  url     = {https://arxiv.org/abs/2608.19652}
}

@article{wang2026medmemorybench,
  title   = {{MedMemoryBench}: Benchmarking Agent Memory in Personalized Healthcare},
  author  = {Wang, Yihao and Xu, Haoran and Gu, Renjie and Ye, Yixuan and Chen, Xinyi and Mu, Xinyu and Gao, Yuan and Guo, Chunxiao and Wei, Peng and Gu, Jinjie and Li, Huan and Chen, Ke and Shou, Lidan},
  journal = {arXiv preprint arXiv:2605.11814},
  year    = {2026},
  url     = {https://arxiv.org/abs/2605.11814}
}

@article{stinard2026clinicalbench,
  title   = {{ClinicalBench}: Stress-Testing Assertion-Aware Retrieval for Cross-Admission Clinical {QA} on {MIMIC-IV}},
  author  = {Stinard, Alex},
  journal = {arXiv preprint arXiv:2605.11143},
  year    = {2026},
  url     = {https://arxiv.org/abs/2605.11143}
}

@inproceedings{snodgrass1985taxonomy,
  title     = {A Taxonomy of Time in Databases},
  author    = {Snodgrass, Richard T. and Ahn, Ilsoo},
  booktitle = {Proceedings of the 1985 ACM SIGMOD International Conference on Management of Data},
  pages     = {236--246},
  year      = {1985},
  doi       = {10.1145/318898.318921}
}

@inproceedings{xu2025amem,
  title = {{A-Mem}: Agentic Memory for {LLM} Agents},
  author    = {Xu, Wujiang and Liang, Zujie and Mei, Kai and Gao, Hang and Tan, Juntao and Zhang, Yongfeng},
  booktitle = {Advances in Neural Information Processing Systems},
  volume    = {38},
  year      = {2025},
  doi       = {10.52202/085713-0593},
  url       = {https://proceedings.neurips.cc/paper_files/paper/2025/hash/19909c36f51abc4856b4560aff3d36d6-Abstract-Conference.html}
}

@inproceedings{chhikara2025mem0,
  title     = {{Mem0}: Building Production-Ready {AI} Agents with Scalable Long-Term Memory},
  author    = {Chhikara, Prateek and Khant, Dev and Aryan, Saket and Singh, Taranjeet and Yadav, Deshraj},
  booktitle = {ECAI 2025},
  pages     = {2993--3000},
  year      = {2025},
  doi       = {10.3233/FAIA251160},
  url       = {https://doi.org/10.3233/FAIA251160}
}

@inproceedings{fang2026lightmem,
  title     = {{LightMem}: Lightweight and Efficient Memory-Augmented Generation},
  author    = {Fang, Jizhan and Deng, Xinle and Xu, Haoming and Jiang, Ziyan and Tang, Yuqi and Xu, Ziwen and Deng, Shumin and Yao, Yunzhi and Wang, Mengru and Qiao, Shuofei and Chen, Huajun and Zhang, Ningyu},
  booktitle = {International Conference on Learning Representations},
  year      = {2026},
  url       = {https://proceedings.iclr.cc/paper_files/paper/2026/hash/a05b72653ec5b473732129829ae04195-Abstract-Conference.html}
}

@article{li2025memos,
  title   = {{MemOS}: A Memory {OS} for {AI} System},
  author  = {Li, Zhiyu and Xi, Chenyang and Li, Chunyu and Chen, Ding and Chen, Boyu and Song, Shichao and Niu, Simin and Wang, Hanyu and Yang, Jiawei and Tang, Chen and Yu, Qingchen and Zhao, Jihao and Wang, Yezhaohui and Liu, Peng and Lin, Zehao and Wang, Pengyuan and Huo, Jiahao and Chen, Tianyi and Chen, Kai and Li, Kehang and Tao, Zhen and Lai, Huayi and Wu, Hao and Tang, Bo and Wang, Zhengren and Fan, Zhaoxin and Zhang, Ningyu and Zhang, Linfeng and Yan, Junchi and Yang, Mingchuan and Xu, Tong and Xu, Wei and Chen, Huajun and Wang, Haofen and Yang, Hongkang and Zhang, Wentao and Xu, Zhi-Qin John and Chen, Siheng and Xiong, Feiyu},
  journal = {arXiv preprint arXiv:2507.03724},
  year    = {2025},
  url     = {https://arxiv.org/abs/2507.03724}
}

@article{zhang2026memrl,
  title   = {{MemRL}: Self-Evolving Agents via Runtime Reinforcement Learning on Episodic Memory},
    author  = {Zhang, Shengtao and Wang, Jiaqian and Zhou, Ruiwen and Liao, Junwei and Feng, Yuchen and Li, Zhuo and Zheng, Yujie and Zhang, Weinan and Wen, Ying and Li, Zhiyu and Xiong, Feiyu and Qi, Yutao and Tang, Bo and Wen, Muning},
  journal = {arXiv preprint arXiv:2601.03192},
  year    = {2026},
  url     = {https://arxiv.org/abs/2601.03192}
}

@inproceedings{yan2026memoryr1,
  title     = {{Memory-R1}: Enhancing Large Language Model Agents to Manage and Utilize Memories via Reinforcement Learning},
  author    = {Yan, Sikuan and Yang, Xiufeng and Huang, Zuchao and Nie, Ercong and Ding, Zifeng and Li, Zonggen and Ma, Xiaowen and Bi, Jinhe and Kersting, Kristian and Pan, Jeff Z. and Schuetze, Hinrich and Tresp, Volker and Ma, Yunpu},
  booktitle = {Proceedings of the 64th Annual Meeting of the Association for Computational Linguistics (Volume 1: Long Papers)},
  pages     = {12805--12825},
  address   = {San Diego, California, United States},
  publisher = {Association for Computational Linguistics},
  year      = {2026},
  doi       = {10.18653/v1/2026.acl-long.583},
  url       = {https://aclanthology.org/2026.acl-long.583/}
}

@article{rasmussen2025zep,
  title   = {Zep: A Temporal Knowledge Graph Architecture for Agent Memory},
  author  = {Rasmussen, Preston and Paliychuk, Pavlo and Beauvais, Travis and Ryan, Jack and Chalef, Daniel},
  journal = {arXiv preprint arXiv:2501.13956},
  year    = {2025},
  url     = {https://arxiv.org/abs/2501.13956}
}

@inproceedings{su2026tsm,
  title     = {Beyond Dialogue Time: Temporal Semantic Memory for Personalized {LLM} Agents},
  author    = {Su, Miao and Guo, Yucan and Hou, Zhongni and Bai, Long and Li, Zixuan and Zhang, Yufei and Yin, Guojun and Lin, Wei and Jin, Xiaolong and Guo, Jiafeng and Cheng, Xueqi},
  booktitle = {Findings of the Association for Computational Linguistics: ACL 2026},
  pages     = {29935--29951},
  address   = {San Diego, California, United States},
  publisher = {Association for Computational Linguistics},
  year      = {2026},
  doi       = {10.18653/v1/2026.findings-acl.1496},
  url       = {https://aclanthology.org/2026.findings-acl.1496/}
}

@inproceedings{banerjee2026apexmem,
  title     = {{APEX-MEM}: Agentic Semi-Structured Memory with Temporal Reasoning for Long-Term Conversational {AI}},
  author    = {Banerjee, Pratyay and Moshtaghi, Masud and Subramanian, Shivashankar and Misra, Amita and Chadha, Ankit},
  booktitle = {Proceedings of the 64th Annual Meeting of the Association for Computational Linguistics (Volume 1: Long Papers)},
  pages     = {16470--16489},
  address   = {San Diego, California, United States},
  publisher = {Association for Computational Linguistics},
  year      = {2026},
  doi       = {10.18653/v1/2026.acl-long.749},
  url       = {https://aclanthology.org/2026.acl-long.749/}
}

@article{yadav2026memstrata,
  title   = {Temporal Validity in Retrieval Memory: Eliminating Stale-Fact Errors for {AI} Agents over Evolving Knowledge},
  author  = {Yadav, Neeraj},
  journal = {arXiv preprint arXiv:2606.26511},
  year    = {2026},
  url     = {https://arxiv.org/abs/2606.26511}
}

@article{wang2026toki,
  title   = {{TOKI}: A Bitemporal Operator Algebra for Contradiction Resolution in {LLM}-Agent Persistent Memory},
  author  = {Wang, Ziming},
  journal = {arXiv preprint arXiv:2606.06240},
  year    = {2026},
  url     = {https://arxiv.org/abs/2606.06240}
}

@article{cui2025timer,
  title   = {{TIMER}: Temporal Instruction Modeling and Evaluation for Longitudinal Clinical Records},
  author  = {Cui, Hejie and Unell, Alyssa and Chen, Bowen and Fries, Jason Alan and Alsentzer, Emily and Koyejo, Sanmi and Shah, Nigam H.},
  journal = {npj Digital Medicine},
  volume  = {8},
  pages   = {577},
  year    = {2025},
  doi     = {10.1038/s41746-025-01965-9},
  url     = {https://doi.org/10.1038/s41746-025-01965-9}
}

@article{xu2026longmedbench,
  title   = {{LongMedBench}: Benchmarking Medical Agents for Long-Horizon Clinical Decision-Making},
  author  = {Xu, Zihan and Chen, Yanzhen and Zhang, Xiaocheng and Fan, Zhiting and Zhai, Weiqi and Xu, Hongxia and Liu, Zuozhu},
  journal = {arXiv preprint arXiv:2607.09322},
  year    = {2026},
  url     = {https://arxiv.org/abs/2607.09322}
}

@article{zhang2026medlocomo,
  title   = {{MedLoCoMo}: A Long-Context Multi-Session Medical Dialogue Benchmark for Large Language Models},
  author  = {Zhang, Zeyu and Wang, Ziqing and Ding, Kaize},
  journal = {arXiv preprint arXiv:2607.22566},
  year    = {2026},
  url     = {https://arxiv.org/abs/2607.22566}
}

@article{wang2026clintracebench,
  title   = {{ClinTraceBench}: Source-Verifiable Longitudinal Clinical Reasoning over {EHR}-Derived Dialogues},
  author  = {Wang, Huimin and Zhao, Zhengyi and Zhao, Yutian},
  journal = {arXiv preprint arXiv:2609.01111},
  year    = {2026},
  url     = {https://arxiv.org/abs/2609.01111}
}

@article{chan2026medcache,
  title   = {{MedCache}: Efficient and Temporally Valid Memory for Longitudinal Clinical Agents},
  author = {Chan, Hei Ting (Una) and Wu, Chenwei and Liu, Xueshen and Zheng, Boyuan and Shen, Liyue and Chen, Jiasi and Mao, Z. Morley},
  journal = {arXiv preprint arXiv:2608.29528},
  year    = {2026},
  url     = {https://arxiv.org/abs/2608.29528}
}

@article{johnson2023mimiciv,
  title   = {MIMIC-IV, a Freely Accessible Electronic Health Record Dataset},
  author  = {Johnson, Alistair E. W. and Bulgarelli, Lucas and Shen, Lu and Gayles, Alvin and Shammout, Ayad and Horng, Steven and Pollard, Tom J. and Hao, Sicheng and Moody, Benjamin and Gow, Brian and Lehman, Li-wei H. and Celi, Leo A. and Mark, Roger G.},
  journal = {Scientific Data},
  volume  = {10},
  number  = {1},
  pages   = {1},
  year    = {2023},
  doi     = {10.1038/s41597-022-01899-x}
}

@inproceedings{wu2024mimicinstr,
  title     = {Instruction Tuning Large Language Models to Understand Electronic Health Records},
  author    = {Wu, Zhenbang and Dadu, Anant and Nalls, Mike and Faghri, Faraz and Sun, Jimeng},
  booktitle = {Advances in Neural Information Processing Systems},
  volume    = {37},
  year      = {2024},
  note      = {Datasets and Benchmarks Track},
  doi       = {10.52202/079017-1737}
}

@inproceedings{asai2024selfrag,
title     = {{Self-RAG}: Learning to Retrieve, Generate, and Critique through Self-Reflection},
author    = {Asai, Akari and Wu, Zeqiu and Wang, Yizhong and Sil, Avirup and Hajishirzi, Hannaneh},
booktitle = {International Conference on Learning Representations},
year      = {2024},
url       = {https://proceedings.iclr.cc/paper_files/paper/2024/hash/25f7be9694d7b32d5cc670927b8091e1-Abstract-Conference.html}
}

@inproceedings{jeong2024adaptiverag,
title     = {{Adaptive-RAG}: Learning to Adapt Retrieval-Augmented Large Language Models through Question Complexity},
author    = {Jeong, Soyeong and Baek, Jinheon and Cho, Sukmin and Hwang, Sung Ju and Park, Jong C.},
booktitle = {Proceedings of the 2024 Conference of the North American Chapter of the Association for Computational Linguistics: Human Language Technologies (Volume 1: Long Papers)},
pages     = {7036--7050},
address   = {Mexico City, Mexico},
publisher = {Association for Computational Linguistics},
year      = {2024},
doi       = {10.18653/v1/2024.naacl-long.389},
url       = {https://aclanthology.org/2024.naacl-long.389/}
}

@article{yang2025qwen3,
  title   = {Qwen3 Technical Report},
  author  = {Yang, An and Li, Anfeng and Yang, Baosong and others},
  journal = {arXiv preprint arXiv:2505.09388},
  year    = {2025},
  url     = {https://arxiv.org/abs/2505.09388}
}

@inproceedings{xiao2023cpack,
  title     = {{C-Pack}: Packed Resources For General Chinese Embeddings},
  author    = {Xiao, Shitao and Liu, Zheng and Zhang, Peitian and Muennighoff, Niklas
               and Lian, Defu and Nie, Jian-Yun},
  booktitle = {Proceedings of the 47th International ACM SIGIR Conference on Research and Development in Information Retrieval},
  pages     = {641--649},
  publisher = {Association for Computing Machinery},
  year      = {2024},
  doi       = {10.1145/3626772.3657878},
  url       = {https://doi.org/10.1145/3626772.3657878}
}

@inproceedings{kwon2023vllm,
  title={Efficient Memory Management for Large Language Model Serving with
         PagedAttention},
  author={Kwon, Woosuk and Li, Zhuohan and Zhuang, Siyuan and Sheng, Ying
          and Zheng, Lianmin and Yu, Cody Hao and Gonzalez, Joseph E.
          and Zhang, Hao and Stoica, Ion},
  booktitle={Proceedings of the 29th Symposium on Operating Systems Principles},
  pages={611--626},
  year={2023},
  doi={10.1145/3600006.3613165}
}

@inproceedings{uddin2026memora,
  title={From Recall to Forgetting: Benchmarking Long-Term Memory for
         Personalized Agents},
  author={Uddin, Md Nayem and Shubham, Kumar and Blanco, Eduardo
          and Baral, Chitta and Wang, Gengyu},
  booktitle={Findings of the Association for Computational Linguistics:
             ACL 2026},
  pages={26814--26841},
  year={2026},
  doi={10.18653/v1/2026.findings-acl.1337}
}

\appendix
\section{Dataset and Evaluation Protocols}
\label{app:dataset_protocols}

MedMemoryBench contains 20 synthetic chronic-disease personas, each with 101 physician--patient sessions. We evaluate all questions from personas 1--5, covering 505 sessions and 496 questions, with no additional sampling or filtering. These comprise 100 entity exact match, 100 temporal localization, 100 inference generation, 99 multiple choice, 50 state update, and 47 multi-hop clinical deduction questions. At each query checkpoint, only sessions observed up to that point are available. MedLoCoMo contains 100 patient timelines and 17,892 QA items. We evaluate a deterministic 320-question subset spanning 12 patients, stratified by timeline length and question scope/type after partitioning patients. The subset contains 96 adversarial, 48 longitudinal-progression, 48 care-plan-rationale, 48 medical-reasoning, 40 cross-admission-comparison, and 40 frequency-pattern questions, including both single-admission and cross-admission cases.

We construct MIMIC-QA from MIMIC-IV using canonical question templates and answer operations adapted from MIMIC-Instr~\citep{wu2024mimicinstr}. Of the 89 published templates across 11 event families, 88 are reproducible in our environment; the chief-complaint template requires MIMIC-IV-Note and is therefore unavailable. Under our cohort and per-patient question budget, 71 templates from eight event families are instantiated: laboratory, input, chart, output, prescription, diagnosis, procedure, and microbiology. The cohort contains 100 patients and 118 admissions sampled from the intersection of hospital admissions and ICU stays using seed 42. All admissions from a patient remain in the same partition. Templates are instantiated for each admission, and round-robin sampling across event families retains 20 questions per patient, yielding 2,000 QA pairs. No LLM is used for question generation or paraphrasing. Patients are split 70/10/20 into train, validation, and test partitions, yielding 1,400/200/400 questions and 85/10/23 admissions. We evaluate the full 400-question test split. Each answer is recomputed from its supporting rows and verified against the recorded answer; all 2,000 items pass this check. Among the 2,000 questions, 301 have Boolean answers, with a 66.5\% Yes rate.

ClinicalBench is evaluated using its full release of 400 questions from 43 patients and 75 referenced admissions. It contains 110 negation, 50 current-state, 50 historical, 40 uncertainty, 40 sequence, 30 family-history, 30 change, 30 duration, and 20 conditional questions. Sequence, change, and duration account for 100 cross-admission questions; the remaining 300 are admission-specific. We use the Hugging Face release \texttt{alexstinard/epikg-clinicalbench} at revision \texttt{e9d70b2a02756de69ea1c4a42d64e5728c833cb6}.

\paragraph{Record length.}
Table~\ref{tab:dataset_context_stats} summarizes the scale of the complete per-person records used in our evaluated cohorts. Token counts are computed with the Qwen3-4B-Instruct-2507 tokenizer and include only the underlying record content supplied to the memory system, excluding prompt scaffolding and formatting introduced by our pipeline. Across datasets, individual records span from 3.4K to 815.0K tokens.

\begin{table}[t]
\centering
\small
\setlength{\tabcolsep}{4pt}
\caption{Per-person context-length statistics for the evaluated cohorts. 
``Records'' denotes sessions for MedMemoryBench and admissions for the 
remaining datasets. Token counts cover each individual's complete record.}
\label{tab:dataset_context_stats}
\begin{tabular}{lrrrrr}
\toprule
Dataset & Persons & Records & Mean tokens & Median & Range \\
\midrule
MedMemoryBench   & 5  & 101.0 & 718.9K & 666.4K & 644.5K--815.0K \\
MedLoCoMo        & 12 & 31.1  & 37.9K  & 37.1K  & 22.6K--61.8K \\
MIMIC-QA (test)  & 20 & 1.1   & 64.7K  & 44.8K  & 9.8K--211.1K \\
ClinicalBench    & 43 & 5.2   & 20.0K  & 15.0K  & 3.4K--73.0K \\
\bottomrule
\end{tabular}
\end{table}

For MedMemoryBench, MedLoCoMo, and ClinicalBench, token counts are computed over the textual record content provided by the corresponding benchmark. MIMIC-QA is constructed from structured MIMIC-IV rows and therefore has no native free-text record representation. For MIMIC-QA, we instead count tokens in the deterministic textual rendering of those rows used by our memory pipeline; the same representation is supplied to every evaluated system. The MedMemoryBench and MedLoCoMo statistics in Table~\ref{tab:dataset_context_stats} are computed over the subsets used in our experiments.

\subsection{Evaluation Scoring}
\label{app:evaluation_scoring}

Evaluation follows benchmark-specific scoring protocols.
Table~\ref{tab:eval_scorers} summarizes the scoring procedure for each benchmark.
All LLM-based evaluation uses the fixed \texttt{gpt-5-2025-08-07} snapshot
with a judge-specific prompt.

\begin{table}[t]
\centering
\small
\setlength{\tabcolsep}{4pt}
\caption{Evaluation scoring by benchmark.}
\label{tab:eval_scorers}
\begin{tabular}{p{0.21\linewidth} p{0.39\linewidth} p{0.30\linewidth}}
\toprule
Benchmark & Question types & Scoring \\
\midrule
MedMemoryBench
& Entity exact match; multiple choice
& Deterministic \\

& Temporal localization; state update; inference generation; multi-hop clinical deduction
& GPT-5 judge \\[2pt]

MedLoCoMo
& All six question types
& Macro deterministic token-level F1 \\[2pt]

MIMIC-QA
& All question types
& Mean deterministic token-level F1 \\[2pt]

ClinicalBench
& All nine question types
& GPT-5 judge against physician-corrected references \\
\bottomrule
\end{tabular}
\end{table}

\subsection{Reader Repeatability and Statistical Controls}
\label{app:reader_repeat}
\label{app:store_gate_controls}

Serving-action comparisons are paired over the same questions, and each cross-dataset ablation row compares conditions evaluated within the same serving pipeline. To measure reader variation independently of memory construction and retrieval, we re-answer byte-identical reference packets within the same evaluation process. On MedMemoryBench, the repeated score changes from 36.3 to 36.9. On MIMIC-QA, repeated answering of identical evidence packets yields 55.34, 55.30, and 55.34 Token-F1, a range of 0.04 points.

Binary correctness differences are evaluated with exact McNemar tests.
For paired Token-F1 comparisons, we use paired-bootstrap 95\% confidence
intervals over question-level score differences and two-sided paired
sign-flip permutation tests. We interpret small answer-level differences alongside the repeat controls and use direct state- or evidence-level metrics when assessing components whose intended effects occur before answer generation.

Results from different answering passes are not treated as interchangeable.
All score differences and paired tests are computed within the corresponding
evaluation pass. Store-gate comparisons are paired against the corresponding
BOTH condition, and write-time ablations are interpreted only within their
reported ablation comparisons.

\section{Implementation Details and Reproducibility}
\label{app:implementation}
\paragraph{Model configuration.}
The two panels in Table~\ref{tab:main_results} identify the answering backbone, Qwen3-4B-Instruct or GPT-5. The local STAM write modules use Qwen3-1.7B. We distinguish memory construction from answer generation because several configurations use different models for these stages. Table~\ref{tab:main_results} reports the headline system-level
configurations. Ablation tables use separate paired evaluation passes and
configuration-specific serving settings; their absolute scores should
therefore not be compared directly with Table~\ref{tab:main_results}.

\paragraph{Baseline implementations.}
We evaluate LightMem, A-Mem, MemOS, Mem0, and MemRL using
the implementations integrated into the MedMemoryBench
evaluation framework. CUPMem uses the authors' released implementation (commit
\texttt{ea7d391}) without modifications to the memory engine or its default
thresholds. Because the clinical benchmarks do not provide oracle-marked
conflict sessions, CUPMem processes the full observed history. We replace
the GPT-4o-mini backbone used in the original CUPMem experiments with the
model used in our evaluation setup, and map the clinical dialogue turns to
CUPMem's conversational input format. Graphiti (Zep) uses self-hosted
Graphiti version 0.29.2. The recorded revisions are \texttt{8d70807} for the MedMemoryBench framework and \texttt{a19ea88} for LightMem. CUPMem was introduced as a targeted prototype for the STALE setting rather
than a general-purpose memory architecture; we therefore treat its results
here as an out-of-domain evaluation of the released method.

\subsection{Full-Context Truncation}
\label{app:full_context_truncation}

Throughout the paper, \emph{Full Context} denotes direct answering from the
temporally ordered raw longitudinal record, without memory construction or
retrieval. When the raw record exceeds the reader-specific input budget, we
remove the oldest record entries first until the input fits.

No truncation is required on MedLoCoMo or ClinicalBench because all evaluated
raw records fit within their respective context limits. MIMIC-QA requires
truncation for a subset of questions. Under Qwen3-4B, the 110K-token evidence
cap affects 179/400 questions (44.8\%); among affected questions, a median of
35.1\% of the raw-record tokens is removed. Under GPT-5, the 280K-token cap
affects 21/400 questions (5.2\%); across the complete 400-question test set,
5.6\% of raw-record tokens are removed.

MedMemoryBench contains substantially longer records. The GPT-5 Full Context
configuration uses a 275{,}388-token evidence cap and truncates the oldest
dialogue turns when necessary.

\begin{table}[t]
\centering
\small
\setlength{\tabcolsep}{4pt}
\caption{Raw-record truncation in the reported Full Context evaluations.
Only final Full Context configurations reported in the main results are included.}
\label{tab:full-context-truncation}
\begin{tabular}{llrr}
\toprule
Benchmark & Reader & Evidence cap & Questions truncated \\
\midrule
MedMemoryBench & GPT-5    & 275K & 329 / 496 (66.3\%) \\
MedLoCoMo      & Qwen3-4B & 100K & 0 / 320 (0\%) \\
MedLoCoMo      & GPT-5    & 100K & 0 / 320 (0\%) \\
MIMIC-QA       & Qwen3-4B & 110K & 179 / 400 (44.8\%) \\
MIMIC-QA       & GPT-5    & 280K & 21 / 400 (5.2\%) \\
ClinicalBench  & Qwen3-4B & 120K & 0 / 400 (0\%) \\
ClinicalBench  & GPT-5    & 120K & 0 / 400 (0\%) \\
\bottomrule
\end{tabular}
\end{table}

\paragraph{Retrieval settings.}
LightMem, A-Mem, MemOS, and Mem0 use retrieval depths of 20, 5, 5, and 5, respectively; MemRL selects 2 memories from 20 candidates.
These five methods follow the MedMemoryBench retrieval configuration.
Graphiti (Zep) retrieves up to 10 items from each of three channels: facts, entities, and episodes.
CUPMem uses internal evidence selection during answering, without an externally specified retrieval depth.
Because memory representations differ, retrieval counts do not determine a common context-token budget.
Table~\ref{tab:baseline_retrieval_settings} therefore reports both selection settings and observed median evidence lengths for the evaluated configurations.

\paragraph{Evidence-token measurement.}
We count the evidence supplied to the reader using the Qwen tokenizer, excluding prompt scaffolding.
Graphiti evidence is measured after its episode truncation to 1,200 characters per episode.
These measurements describe observed context lengths rather than imposed token limits.
The GPT-5 MedMemoryBench estimates for LightMem, A-Mem, MemOS, and Mem0 cover the 199 of 496 questions with saved evidence blocks; the corresponding MemRL blocks were not saved, and the CUPMem estimate covers one persona.
The rendered-context counts used in the budget-matched analysis follow a different counting convention and are not directly comparable to these evidence-only counts.

\begin{table}[t]
\centering
\small
\setlength{\tabcolsep}{3pt}
\renewcommand{\arraystretch}{1.05}
\begin{tabular}{llrrrr}
\toprule
Method & Selection & MMB & MedLoCoMo & MIMIC-QA & ClinicalBench \\
\midrule
\multicolumn{6}{c}{\textit{Qwen3-4B-Instruct}} \\
\midrule
LightMem       & 20         & 1.1  & 0.8 & 1.3  & 1.1 \\
A-Mem          & 5          & 23.6 & 8.0 & 18.0 & 8.1 \\
MemOS          & 5          & 0.4  & 0.4 & 1.0  & 0.7 \\
Mem0           & 5          & 0.1  & 0.1 & 0.1  & 0.1 \\
MemRL          & 2 of 20    & 0.5  & 0.6 & 0.7  & 0.7 \\
CUPMem         & Internal   & 0.3  & 0.2 & 0.1  & 0.2 \\
Graphiti (Zep) & 10/channel & 4.1  & 3.9 & 3.7  & 2.4 \\
\midrule
\multicolumn{6}{c}{\textit{GPT-5}} \\
\midrule
LightMem       & 20         & 1.0$^{a}$ & 0.7 & 1.3       & 1.1 \\
A-Mem          & 5          & 21.8$^{a}$& 7.9 & 18.1$^{c}$& 8.2 \\
MemOS          & 5          & 0.4$^{a}$ & 0.3 & 1.1$^{c}$ & 0.5 \\
Mem0           & 5          & 0.2$^{a}$ & 0.1 & 0.2       & 0.1 \\
MemRL          & 2 of 20    & NR        & 0.6 & 0.7       & 0.7 \\
CUPMem         & Internal   & 0.1$^{b}$ & 0.0 & NA        & 0.0 \\
Graphiti (Zep) & 10/channel & 4.8       & 4.4 & 3.8       & 2.6 \\
\bottomrule
\end{tabular}
\caption{
Baseline selection settings and median evidence tokens per question (thousands).
MMB denotes MedMemoryBench.
$^{a}$199 logged questions; $^{b}$one persona;
$^{c}$Qwen3-4B memories read by GPT-5.
NR: not recorded; NA: unavailable.
CUPMem reports only evidence passing its premise check, so 0.0 does not imply an empty store.
}
\label{tab:baseline_retrieval_settings}
\end{table}

\paragraph{Answering protocols and adaptations.} In the local MIMIC-QA panel, the LightMem, A-Mem, MemOS, Mem0, and MemRL scores are obtained by passing their retrieved memories to a shared reader, whereas Graphiti uses its own answering step. In the GPT-5 MIMIC-QA panel, A-Mem and MemOS use memories constructed with Qwen3-4B and answered with GPT-5. On ClinicalBench, the five MedMemoryBench-integrated baselines retain the MedMemoryBench prompts: each note is presented as a medical dialogue record, and the answer prompt is the MedMemoryBench default. Graphiti and CUP-Mem answer with their own prompts.

\paragraph{Evidence visibility.}
For the GPT-5 STAM configuration on ClinicalBench,
retrieval is restricted to the source notes authorized
for each question. State labels respect the same
visibility boundary: a memory is marked historical only
when its transition is supported by visible evidence.
A successor in an unavailable future note alone does not
change the memory's visible state.

\subsection{Write-Module Fine-Tuning}
\label{app:write_finetuning}
\paragraph{Trainable components.} We fine-tune the extractor, relation linker, and updater using separate LoRA adapters on Qwen3-1.7B. MIMIC-QA does not use an extractor, so only its relation linker and updater are adapted.  We train a separate adapter set for each benchmark and, for ClinicalBench, for each evaluation fold. 

\paragraph{Data separation.}
Training and validation are separated from evaluation
at the patient or persona level, as summarized in
Table~\ref{tab:write_finetuning_splits}. For MedLoCoMo,
patient~16957952 provides validation data. For MIMIC-QA,
teacher examples come from ten admissions belonging to
ten patients in the benchmark's training split. A seeded
shuffle holds out admission~23790113 from
patient~11348441 for validation; the remaining nine
admissions provide training examples for both adapters.
None of these ten patients belongs to the benchmark's
validation or test split.

ClinicalBench uses five-fold evaluation. Training,
validation, and evaluation patients are disjoint within
each fold, and predictions on the held-out patients are
combined across folds for the 400-question evaluation.

\begin{table}[t]
\centering
\small
\setlength{\tabcolsep}{4pt}
\renewcommand{\arraystretch}{1.08}
\begin{tabularx}{\linewidth}{lXXX}
\toprule
Benchmark & Training & Validation & Evaluation \\
\midrule
MedMemoryBench
& Personas 6, 8--12
& Persona 7
& Personas 1--5 \\

MedLoCoMo
& 5 patients
& 1 patient
& 12 patients; 320 questions \\

MIMIC-QA
& 9 admissions from 9 patients
& 1 admission from 1 patient
& 20 patients \\

ClinicalBench
& 30--31 patients per fold
& 4 patients per fold
& 8--9 patients per fold; 400 questions overall \\
\bottomrule
\end{tabularx}
\caption{Data partitions for write-module fine-tuning.
MIMIC-QA adapter validation is an internal holdout from
the benchmark's training split. ClinicalBench combines
predictions from five held-out folds.}
\label{tab:write_finetuning_splits}
\end{table}

\paragraph{Optimization.}
We minimize cross-entropy on teacher-generated response
tokens, masking the input prompt from the loss. LoRA
uses rank~32, scaling factor~64, and dropout~0.05,
and is applied to all attention and MLP projection
layers. The learning rate is $10^{-4}$, with a warm-up
fraction of $0.03$ and gradient clipping at~1.0.
The effective batch size is~8, using a microbatch size
of~1 and eight gradient-accumulation steps. Training
runs for at most two epochs. The random seed is
20260912, except for MIMIC-QA, which uses 20260908.
For the MedLoCoMo extractor, the end-of-sequence token
receives three times the standard loss weight.

\section{Additional Experimental Analysis}
\label{app:additional_experiments}

This section specifies the evidence definitions and retrieval metrics used
across benchmarks, including the treatment of incomplete evidence annotations.
We examine retrieval coverage, context budget, and downstream answer quality
under controlled serving configurations.

\subsection{Retrieval Diagnostics}
\label{app:retrieval_diagnostics}

\subsubsection{Retrieval Coverage and Context Budget}

Table~\ref{tab:retrieval_efficiency} reports retrieval coverage under the
configurations used for downstream QA. Because methods provide different
amounts of context to the answering model, these results should not be
interpreted as matched-budget retrieval rankings.

\begin{table*}[t]
\centering
\small
\setlength{\tabcolsep}{6pt}
\renewcommand{\arraystretch}{1.05}
\begin{tabular}{lrrrrrr}
\toprule
& \multicolumn{2}{c}{MedMemoryBench}
& \multicolumn{2}{c}{MedLoCoMo}
& \multicolumn{2}{c}{MIMIC-QA} \\
\cmidrule(lr){2-3}
\cmidrule(lr){4-5}
\cmidrule(lr){6-7}
Method
& Tokens/q & Recall
& Tokens/q & Recall
& Tokens/q & Recall \\
\midrule
Flat RAG
& 1.9K & 0.536
& 1.3K & 0.560
& 2.3K & 0.608 \\

LightMem
& 1.3K & 0.727
& 0.8K & 0.624
& 1.3K & 0.515$^{\ddagger}$ \\

MemOS
& 0.4K & 0.657
& 0.4K & 0.543
& 1.0K & 0.338$^{\ddagger}$ \\

Mem0
& 0.1K & 0.618
& 0.1K & 0.543
& 0.1K & 0.500$^{\ddagger}$ \\

Ours
& 4.2K & 0.802
& 4.1K & 0.822
& 2.3K & 0.959 \\
\bottomrule
\end{tabular}
\caption{
Retrieval coverage under the configurations used for downstream QA.
Because methods provide different amounts of context to the answering model,
these results characterize their evaluation configurations rather than
matched-budget retrieval comparisons.
Recall uses dataset-specific gold evidence annotations and is not comparable
across datasets. ClinicalBench has no evidence-level gold annotations.
}
\label{tab:retrieval_efficiency}
\end{table*}

\subsubsection{Evidence Definitions}
\label{app:retrieval_precision}

Evidence annotations differ across the three benchmarks for which retrieval can be evaluated. On MedMemoryBench, a cited source key point is considered retrieved when at least one served memory has cosine similarity at least $\tau=0.75$ under BGE-small. Because the gold consists of benchmark key points, some of which may never have been written into the memory store by the extractor agent, this metric reflects both store coverage and read-time retrieval. Of the 496 evaluated questions, 401 contain resolvable cited evidence. On MedLoCoMo, gold evidence consists of store atoms whose normalized source spans fall within cited dialogue turns. Depending on the evaluated store, 128--143 of the 320 questions contain resolvable turn-level gold. On MIMIC-QA, 336 of 400 questions have gold defined as a specific set of source rows; an additional 43 questions are satisfied by any one of several sufficient rows, and 21 negative-existence questions have no positive gold row by construction. The headline retrieval metric uses the 336 questions with definite-set gold. ClinicalBench does not provide evidence-level annotations and is therefore excluded from retrieval-coverage analysis.

Because the benchmarks annotate only a small number of gold evidence items relative
to the number of memories served, raw retrieval precision is strongly constrained by
the serving budget; we therefore emphasize evidence recall together with served
context size rather than precision alone.

\subsubsection{Evidence Coverage and Full-Context Performance}
\label{app:full-context-coverage}
MIMIC-QA provides a high-coverage setting in which to compare selective
serving with raw Full Context. The downstream STAM configuration attains
0.959 gold-evidence recall. Selective serving obtains 87.22 Token-F1
compared with 85.59 for raw Full Context ($\Delta = +1.63$, $p=.18$).
Empty responses are re-evaluated under the same fixed answering
configuration. Thus, when retrieval coverage is high, a compact memory
packet can preserve strong-reader answer quality.

MedMemoryBench provides the contrasting regime. Its downstream evidence recall
is 0.802, and exposing more of the stored memory to GPT-5 substantially improves
answer quality. Together, these results identify evidence selection as an
important bottleneck between the maintained memory and downstream answering.

\subsubsection{GPT-5 Serving-Budget Analysis}
\label{app:gpt5-serving-budget}

The default STAM configurations use bounded evidence serving, so their
downstream scores need not reflect the information ceiling of the maintained
memory. We therefore evaluate the same GPT-5 reader while progressively
relaxing the serving restriction. These serving-budget analyses are diagnostic and are not used as headline
results.

\begin{table*}[t]
\centering
\small
\setlength{\tabcolsep}{4pt}

\begin{minipage}[t]{0.46\linewidth}
\centering
\textit{(a) MedMemoryBench}\\[3pt]
\begin{tabular}{lr}
\toprule
Condition & Macro Acc. \\
\midrule
Raw Full Context                & 63.17 \\
STAM, store-aware $k=120$       & 52.90 \\
STAM, store-aware $k=240$       & 54.47 \\
STAM, flat $k=240$              & 55.50 \\
STAM, flat $k=480$              & 57.40 \\
STAM, flat $k=960$              & 58.61 \\
STAM, all memories              & 63.84 \\
\bottomrule
\end{tabular}
\end{minipage}
\hfill
\begin{minipage}[t]{0.50\linewidth}
\centering
\textit{(b) MedLoCoMo}\\[3pt]
\begin{tabular}{lrr}
\toprule
Condition & mF1 & Words \\
\midrule
Raw Full Context               & 42.53 & 25,157 \\
STAM, $k=60$                   & 36.88 & 423 \\
STAM, $k=240$                  & 39.29 & 1,612 \\
STAM, 4,500-word budget        & 40.75 & 4,494 \\
STAM, all memories             & 38.80 & 10,553 \\
\bottomrule
\end{tabular}
\end{minipage}

\caption{
GPT-5 serving-budget diagnostics.
(a) MedMemoryBench macro accuracy as the serving restriction is relaxed.
(b) MedLoCoMo type-macro F1 and median served words.
Larger-budget configurations are diagnostic analyses and are not used as
headline results.
}
\label{tab:gpt5-serving-budget}
\end{table*}

\paragraph{MedMemoryBench.}
Table~\ref{tab:gpt5-serving-budget}(a) reports macro accuracy, matching the
metric used in Table~\ref{tab:main_results}. Raw Full Context is truncated
oldest-first for 329 of 496 questions.

Serving all maintained STAM memories raises macro accuracy from 52.90 under
the default configuration to 63.84, compared with 63.17 for raw Full Context.

\paragraph{MedLoCoMo.}
MedLoCoMo provides a complementary setting in which the complete raw record
fits within the reader context. All STAM conditions below use the same
maintained memory store.

Increasing the ranked serving budget from the default $k=60$ configuration
to $k=240$ improves type-macro F1 by 2.41 points ($p=.018$), while the
4,500-word ranked condition improves it by 3.87 points ($p=.001$).
The latter reaches 40.75 compared with 42.53 for raw Full Context while
serving approximately $5.6\times$ less text; the paired difference is not
statistically significant ($p=.232$).

Serving all maintained memories in record order reaches 38.80, remaining
3.73 points below the raw record ($p=.009$). Ranked serving therefore
performs better than exposing the complete maintained store while using
substantially less context.

\subsection{Evaluation at Matched Context Lengths}
\label{app:context-matched}

\paragraph{Evaluation protocol.}
Memory methods can differ substantially in the amount
of evidence supplied to the reader, so equal retrieval counts do not imply
comparable context lengths. We therefore compare methods at approximately
matched median rendered context lengths. For each comparator configuration,
we adjust STAM's serving configuration to target the comparator's median
context and retain rows with a STAM-to-comparator ratio between $0.80$ and
$1.25$. Matched-context STAM arms use the same evidence-rendering convention
as the corresponding headline configuration; only the serving budget is
adjusted to target the comparator's median rendered context length.
Standard rows retain the comparator's evaluated configuration; rows labeled
``lifted'' evaluate additional comparator configurations with expanded
context. Both methods use the benchmark harness's reader protocol, with a
separate system turn, a \texttt{[Retrieved Memories]} evidence header,
temperature~1, and a 10{,}000-token completion cap. Throughout this section,
$p$ denotes the paired-test $p$-value under the null hypothesis of no
performance difference; we use $p<0.05$ as the significance threshold.
A nonsignificant result does not establish equivalence. 

\paragraph{Matched-context results.}
Table~\ref{tab:context-matched} reports the primary comparison.
Among the 17 rows with paired tests, STAM has six statistically
significant improvements: over MemRL on MedMemoryBench and MedLoCoMo,
and over Flat RAG, LightMem, MemOS, and MemRL on MIMIC-QA.

\begin{table*}[t]
\centering
\scriptsize
\setlength{\tabcolsep}{3.5pt}
\renewcommand{\arraystretch}{1.08}
\begin{tabular}{llrrrrr}
\toprule
Dataset &
Comparator (ctx.) &
\shortstack{STAM\\ctx.} &
Ratio &
STAM &
Comparator &
$\Delta$ \hspace{3pt} ($p$) \\
\midrule

\multirow{7}{*}{MedMemoryBench}
& LightMem (1,312)          & 1,318 & 1.00 & 39.87 & 39.90 & $-0.03$ \; (.95) \\
& Flat RAG (1,898)          & 1,892 & 1.00 & 37.36 & 39.10 & $-1.74$ \; (.46) \\
& MemOS (449)               & 444   & 0.99 & 30.19 & 32.59 & $-2.40$ \; (.30) \\
& MemRL (596)               & 564   & 0.95 & 29.30 & 22.72 & $+6.58$ \; (.007) \\
& Mem0 (134)                & 157   & 1.17 & 24.75 & 22.62 & $+2.13$ \; (.36) \\
& LightMem lifted (4,505)   & 4,652 & 1.03 & 39.35 & 41.57 & $-2.22$ \; (.28) \\
& Flat RAG lifted (4,594)   & 4,652 & 1.01 & 39.35 & 42.36 & $-3.01$ \; (.21) \\
\midrule

\multirow{5}{*}{MedLoCoMo}
& Flat RAG (1,314)          & 1,353 & 1.03 & 35.73 & 32.23 & $+3.50$ \; (.08) \\
& LightMem (1,026)          & 1,007 & 0.98 & 34.52 & 33.23 & $+1.29$ \; (.48) \\
& MemOS (442)               & 473   & 1.07 & 31.25 & 27.54 & $+3.71$ \; (.07) \\
& MemRL (701)               & 676   & 0.97 & 31.53 & 22.70 & $+8.83$ \; ($<.001$) \\
& Mem0 (105)                & 131   & 1.25 & 28.00 & 28.25 & $-0.26$ \; (.88) \\
\midrule

\multirow{4}{*}{MIMIC-QA}
& Flat RAG (2,282)          & 2,324 & 1.02 & 52.57 & 45.42 & $+7.15$ \; ($<.001$) \\
& LightMem (1,632)          & 1,620 & 0.99 & 51.71 & 22.24 & $+29.47$ \; ($<.001$) \\
& MemOS (1,086)             & 1,168 & 1.08 & 52.67 & 19.90 & $+32.77$ \; ($<.001$) \\
& MemRL (733)               & 734   & 1.00 & 49.19 & 22.15 & $+27.04$ \; ($<.001$) \\
\midrule

\multirow{3}{*}{ClinicalBench}
& LightMem (1,100)          & 890 & 0.81 & 50.00 & 50.50 & $-0.50$ \; (--)\textsuperscript{\dag} \\
& Flat RAG (720)            & 890 & 1.24 & 50.00 & 47.00 & $+3.00$ \; (.28) \\
& MemOS (736)               & 890 & 1.21 & 50.00 & 46.80 & $+3.20$ \; (--)\textsuperscript{\dag} \\
\bottomrule
\end{tabular}

\caption{
Context-matched evaluation.
Context values are median rendered reader tokens, and Ratio is STAM context divided by comparator context.
All arms are answered by the same Qwen3-4B reader under the benchmark harness's reader settings.
MedMemoryBench reports macro accuracy, MedLoCoMo reports type-macro
token-F1, MIMIC-QA reports mean Token-F1, and ClinicalBench reports
pooled accuracy.
Rows labeled ``lifted'' use expanded comparator context.
\textsuperscript{\dag}Per-question comparator outputs are unavailable, so no paired significance test is reported.
}

\label{tab:context-matched}
\end{table*}

\paragraph{ClinicalBench.}
At 890 median rendered tokens, STAM obtains 50.0 accuracy compared with
47.0 for Flat RAG at 720 tokens ($p=0.285$) and 50.25 for the flat
retrieval control at 772 tokens ($p=1.000$). Full Context obtains 51.0
while serving 15{,}619 tokens, compared with STAM's 50.0 using
approximately $6\%$ of that context ($p=0.741$). In contrast, removing
retrieval reduces accuracy to 31.5 ($p<0.001$). These comparisons use
exact McNemar tests over the 400 questions.

\paragraph{MedLoCoMo serving-budget diagnostic.}
Increasing the serving budget also improves the categories most directly tied
to longitudinal reasoning: longitudinal-progression F1 increases from 18.8 to
24.8 and cross-admission-comparison F1 from 10.0 to 17.5.
For reference, raw-record Full Context obtains 28.9 and 15.0 on these two
categories, respectively. Thus, the expanded STAM configuration remains below
raw Full Context on longitudinal progression but exceeds it on cross-admission
comparison. These category-level results are reported as serving-budget diagnostics
rather than headline comparisons.

\paragraph{Comparisons outside the matching range.}
A-Mem and CUPMem fall outside the $0.80$--$1.25\times$ context-matching range.
Table~\ref{tab:context-extra} reports these results separately, and they
are excluded from the matched-context significance summary.

\begin{table}[t]
\centering
\small
\setlength{\tabcolsep}{4pt}
\begin{tabular}{llrrrr}
\toprule
Dataset &
Comparator &
Ratio &
STAM &
Cmp. &
$\Delta$ \; ($p$) \\
\midrule
MMB
& A-Mem  & 0.39 & 35.16 & 33.43 & $+1.73$ \; (.51) \\
& CUPMem & 0.42 & 20.72 & 5.37  & $\mathbf{+15.35}$ \; ($\mathbf{<.001}$) \\
\midrule
MedLoCoMo
& A-Mem  & 0.49 & 34.20 & 36.60 & $-2.40$ \; (.41) \\
& CUPMem & 0.56 & 27.12 & 18.30 & $\mathbf{+8.82}$ \; ($\mathbf{<.001}$) \\
\midrule
MIMIC-QA
& A-Mem  & 0.25 & 54.67 & 42.30 & $\mathbf{+12.37}$ \; ($\mathbf{<.001}$) \\
& CUPMem & 2.12 & 43.96 & 5.92  & $\mathbf{+38.04}$ \; ($\mathbf{<.001}$) \\
\bottomrule
\end{tabular}

\caption{
Comparisons outside the context-matching range.
Ratio denotes STAM context divided by comparator context.
These rows are excluded from the matched-context significance summary.
}
\label{tab:context-extra}
\end{table}

\paragraph{Judge variability.}
On MedMemoryBench, repeated judge evaluations vary by approximately
1--2 macro-accuracy points. The STAM result at the 1{,}312-token
reference context is averaged over three fresh judge evaluations,
whereas the expanded-context rows use one judge evaluation.

\subsection{Retrieval Depth and Evidence Rendering}
\label{app:retrieval-depth}

We examine sensitivity to retrieval depth and evidence rendering under the
local Qwen3-4B-Instruct reader, holding the memory store and reader fixed
within each sweep.

\paragraph{MedMemoryBench.}
Increasing retrieval depth raises evidence recall with diminishing returns
(Table~\ref{tab:retrieval-depth-combined}a): at $\tau=0.75$, recall increases
from 0.709 at $k=30$ to 0.802 at $k=120$ and 0.849 at $k=240$.
Answer quality changes modestly between $k=60$ and $k=120$
(36.97 and 37.86), while the lower score at $k=240$ does not reach
statistical significance relative to $k=120$
(exact McNemar $p=.054$). Retrieval is rerun independently at each depth
rather than approximated by truncating a larger packet.

\paragraph{MedLoCoMo.}
Table~\ref{tab:retrieval-depth-combined}b separates retrieval depth from
evidence rendering. Rendering the source-span context of the same $k=60$
retrieved memories leaves recall unchanged at 0.822 but increases
type-macro F1 from 34.93 to 36.13. Increasing retrieval depth to $k=120$
at a similar context cost raises recall to 0.863 but reaches 34.69 F1.
At a similar context budget, this comparison suggests that allocating
additional context to richer source evidence can be more effective than
retrieving additional memory units, motivating the source-span rendering
used in our MedLoCoMo configuration.
\begin{table*}[t]
\centering
\scriptsize
\setlength{\tabcolsep}{3pt}

\begin{minipage}[t]{0.47\linewidth}
\centering
\textit{(a) MedMemoryBench}\\[3pt]
\resizebox{\linewidth}{!}{
\begin{tabular}{rrrrrr}
\toprule
$k$ & Tokens/q & R@0.70 & R@0.75 & R@0.80 & Macro \\
\midrule
30  & 1,047 & 0.845 & 0.709 & 0.491 & 33.65 \\
60  & 2,122 & 0.892 & 0.760 & 0.531 & 36.97 \\
120 & 4,248 & 0.918 & 0.802 & 0.565 & 37.86 \\
240 & 8,681 & 0.957 & 0.849 & 0.594 & 35.17 \\
\bottomrule
\end{tabular}
}
\end{minipage}
\hfill
\begin{minipage}[t]{0.51\linewidth}
\centering
\textit{(b) MedLoCoMo}\\[3pt]
\resizebox{\linewidth}{!}{
\begin{tabular}{lrrrrr}
\toprule
Setting & Tokens/q & R@0.70 & R@0.75 & R@0.80 & Type-macro F1 \\
\midrule
$k=60$           & 2,422  & 0.948 & 0.822 & 0.591 & 34.93 \\
$k=60$ + span    & 4,111  & 0.948 & 0.822 & 0.591 & 36.13 \\
$k=120$          & 4,668  & 0.960 & 0.863 & 0.642 & 34.69 \\
$k=240$          & 9,128  & 0.971 & 0.896 & 0.689 & 34.93 \\
$k=\mathrm{all}$ & 15,828 & 0.976 & 0.907 & 0.721 & 35.74 \\
$k=240$ + span   & 15,731 & 0.971 & 0.896 & 0.689 & 36.37 \\
\bottomrule
\end{tabular}
}
\end{minipage}

\caption{Retrieval-depth and evidence-rendering diagnostics with the local
Qwen3-4B-Instruct reader. MedMemoryBench uses independent re-retrieval at
each depth. MedLoCoMo span variants retain the same retrieved memories while
additionally providing their source-span context. MedLoCoMo is scored in
type-macro Token-F1; the $k=60+$span configuration corresponds to the
headline setting.}
\label{tab:retrieval-depth-combined}
\end{table*}

\section{Additional Analysis of Store-Aware Serving}
\label{app:store_gate}

\subsection{Paired Store-Gate Evaluation}
\label{app:store_gate_paired}
We evaluate selective HISTORY serving by varying $\lambda$ while holding
the underlying memory store, retrieval packets, and answering configuration
fixed. For each setting, we report the fraction of queries served without
HISTORY and the paired change in answer quality.

Table~\ref{tab:store_gate} reports the selected local-reader gate
configuration for each benchmark.

\begin{table}[t]
\centering
\small
\setlength{\tabcolsep}{4pt}

\begin{tabular}{lrrr}
\toprule
Dataset & $\lambda$ & HISTORY skip & Quality $\Delta$ ($p$) \\
\midrule
MedMemoryBench & 0.01 & 26.2\% & $-0.67$ ($p=.289$) \\
MedLoCoMo      & 0.5  & 51.6\% & $-0.11$ ($p=.469$) \\
MIMIC-QA       & 0.2  & 79.0\% & $-0.70$ ($p=.147$) \\
ClinicalBench  & 0.5  & 1.0\%  & $0.00$ ($p=1.000$) \\
\bottomrule
\end{tabular}

\caption{
Selective HISTORY serving under the published gate configuration.
The gate chooses between the original ACTIVE+HISTORY packet and the same
packet with HISTORY removed. HISTORY skip is the fraction of queries
served without HISTORY. Quality differences are paired against the
corresponding BOTH condition.
}
\label{tab:store_gate}
\end{table}

\subsection{Historical-Memory Dependence}
\label{app:history_requirement}
\label{app:history_by_type}

We further analyze the matched-volume refill control at the question level.
Unlike the Store Gate, which omits HISTORY without refilling the packet, this
analysis replaces the freed context with additional ACTIVE memories. It
therefore tests whether historical evidence remains useful when the ACTIVE-only
condition receives comparable serving capacity.

A question is \emph{History critical} when the A+H answer is correct and the
refilled ACTIVE-only answer is incorrect, and \emph{History harmful} when the
reverse holds. \emph{Active sufficient} includes all questions answered
correctly from the refilled ACTIVE-only condition, including the History-harmful
subset. \emph{Neither correct} denotes questions answered incorrectly under
both conditions.

\begin{table}[t]
\centering
\small
\setlength{\tabcolsep}{3.6pt}
\renewcommand{\arraystretch}{1.05}
\caption{
Question-level outcomes under A+H and refilled ACTIVE-only serving.
A question is correct under the judge verdict for MedMemoryBench and
ClinicalBench, or at token-F1 $\geq 0.5$ for MedLoCoMo and MIMIC-QA.
History harmful is a subset of Active sufficient, so the four columns do
not form a disjoint partition.
}
\label{tab:store_composition}
\begin{tabular}{llrrrr}
\toprule
System & Dataset &
Active sufficient &
History critical &
History harmful &
Neither correct \\
\midrule
\multirow{4}{*}{1.7B write / 4B reader}
& MMB & 35.7\% & 4.8\% & 2.8\% & 59.5\% \\
& MLC & 43.4\% & 0.6\% & 0.3\% & 55.9\% \\
& MQA & 56.2\% & 4.2\% & 3.0\% & 39.5\% \\
& CB  & 50.2\% & 5.2\% & 5.2\% & 44.5\% \\
\midrule
\multirow{4}{*}{GPT-5 write / GPT-5 reader}
& MMB & 47.4\% & 11.9\% & 7.5\% & 40.7\% \\
& MLC & 43.8\% & 4.1\% & 5.6\% & 52.2\% \\
& MQA & 56.2\% & 33.5\% & 2.5\% & 10.2\% \\
& CB  & 50.8\% & 2.5\% & 3.8\% & 46.8\% \\
\bottomrule
\end{tabular}
\end{table}

The question-level analysis shows substantial heterogeneity in dependence on
historical evidence. Under GPT-5, history dependence is particularly pronounced
for several MIMIC-QA temporal queries and remains visible on MedMemoryBench
temporal-localization and state-update questions. The corresponding rates are
much smaller for the local system, consistent with differences in the maintained
stores. These results complement the aggregate historical-access ablation by
showing that the value of HISTORY is concentrated in particular questions and
query types.

\begin{table}[t]
\centering
\small
\setlength{\tabcolsep}{5pt}
\renewcommand{\arraystretch}{1.05}
\caption{
History-critical rates for selected query types under the refilled design:
the fraction of questions answered correctly with A+H but not with refilled
ACTIVE-only serving.
}
\label{tab:history_by_type}
\begin{tabular}{llrrr}
\toprule
Dataset & Query type & $n$ &
1.7B write / 4B reader &
GPT-5 \\
\midrule
MQA & first         & 37 & 2.7\% & 45.9\% \\
MQA & minimum       & 15 & 0.0\% & 60.0\% \\
MQA & maximum       & 11 & 0.0\% & 45.5\% \\
MQA & at\_timestamp & 86 & 4.7\% & 45.3\% \\
MQA & top\_n        & 43 & 0.0\% & 0.0\% \\
\midrule
MMB & temporal localization & 100 & 5.0\% & 15.0\% \\
MMB & state update          & 50  & 8.0\% & 18.0\% \\
\bottomrule
\end{tabular}
\end{table}

\subsection{Store-Aware Retrieval and Rendering Controls}
\label{app:store_aware_retrieval_controls}
\label{app:rendering_2x2}
\label{app:read_diagnostics}

We examine retrieval allocation and evidence rendering separately from the underlying \textsc{Active}/\textsc{History} memory representation. These experiments assess whether allocating retrieval capacity by store or exposing store identity to the reader improves downstream answer quality.

\paragraph{Separate-store versus joint retrieval.}
Table~\ref{tab:joint_ranking_ablation} compares retrieval configurations
using the same underlying memory pool and total retrieval budget.
On MedMemoryBench, the score difference is not statistically significant.
On MIMIC-QA, joint ranking is numerically higher by 0.90 Token-F1
($p=0.52$), but this condition also uses neutral evidence rendering, so
the difference cannot be attributed to retrieval allocation alone.

\begin{table}[t]
\centering
\small
\setlength{\tabcolsep}{5pt}
\renewcommand{\arraystretch}{1.05}
\caption{
Retrieval-configuration comparisons using the same underlying memories
and total retrieval budget. MedMemoryBench reports macro accuracy and
MIMIC-QA reports mean Token-F1. The MIMIC-QA joint-ranking condition
also uses neutral evidence rendering; Table~\ref{tab:mimic_rendering_2x2}
separates retrieval allocation from label rendering.
}
\label{tab:joint_ranking_ablation}
\begin{tabular}{llrr}
\toprule
Dataset & Retrieval & Evidence statistic & Score \\
\midrule
\multirow{2}{*}{MedMemoryBench}
& Store-aware  & History/pkt 44.8\% & 36.3 \\
& Joint ranking & History/pkt 52.3\% & 35.1 \\
\midrule
\multirow{2}{*}{MIMIC-QA}
& Store-aware  & Gold served 97.4\% & 55.34 \\
& Joint ranking & Gold served 98.7\% & 56.23 \\
\bottomrule
\end{tabular}
\end{table}

\paragraph{Retrieval allocation versus store-label rendering.}
We conduct a $2\times2$ analysis on MIMIC-QA, evaluating store-aware and joint-ranking packets with and without explicit \textsc{Current}/\textsc{Historical} labels. This separates changes in the retrieved evidence from changes in how store identity is presented. The reference scores differ from those in Table~\ref{tab:joint_ranking_ablation}, so score differences are interpreted within each experiment.

\begin{table}[t]
\centering
\small
\setlength{\tabcolsep}{3.8pt}
\renewcommand{\arraystretch}{1.05}
\caption{
MIMIC-QA analysis separating retrieval allocation from store-label
rendering. Scores are mean Token-F1 over 400 test questions.
$\Delta$ is the paired mean difference; 95\% confidence intervals are
obtained by paired bootstrap and $p$-values by a two-sided paired
sign-flip permutation test.
}
\label{tab:mimic_rendering_2x2}
\begin{tabular}{lrrrrr}
\toprule
Comparison & From & To & $\Delta$ & 95\% CI & $p$ \\
\midrule
Remove labels, store-aware
& 54.65 & 55.26 & +0.61 & $[-1.8,+3.0]$ & .64 \\

Add labels, joint ranking
& 56.80 & 55.54 & -1.26 & $[-3.9,+1.3]$ & .35 \\

Ranking change, both unlabelled
& 55.26 & 56.80 & +1.54 & $[-0.4,+3.5]$ & .13 \\

Ranking change, both labelled
& 54.65 & 55.54 & +0.89 & $[-1.3,+3.1]$ & .43 \\

Labelled Full $\rightarrow$ unlabelled merged
& 54.65 & 56.80 & +2.15 & $[-0.2,+4.5]$ & .08 \\
\bottomrule
\end{tabular}
\end{table}

None of the isolated retrieval-allocation or label-rendering contrasts
is statistically significant. Removing labels from store-aware packets
changes Token-F1 by $+0.61$ ($p=0.64$), while adding labels to
joint-ranking packets changes it by $-1.26$ ($p=0.35$). Changing from
labelled store-aware retrieval to unlabelled joint ranking yields the
largest numerical difference, $+2.15$ Token-F1, but this difference is
also nonsignificant ($p=0.08$). Thus, these results do not provide clear
evidence that either separate retrieval allocation or explicit store
labels independently improves downstream QA.

\section{Write-Time State-Maintenance Diagnostics}
\label{app:write_state_maintenance}
Table~\ref{tab:ablation_write} reports the main write-time component
ablations. On MedMemoryBench, we evaluate supersession-pair recall and false
archival. On MIMIC-QA, we report the fraction of memories archived, the
number of gold-evidence memories placed in \textsc{History}, and downstream
Token-F1. These diagnostics characterize how the write path organizes
longitudinal state, while downstream QA measures the effect of that state on
answer generation.

\subsection{Scope of Component Ablations}
\label{app:ablation_dataset_scope}

The write-time component ablations cover MedMemoryBench and MIMIC-QA. They complement the historical-access and routing analyses in
Section~\ref{sec:abl_history} and Appendix~\ref{app:store_gate}. Evidence placement and answer-level dependence on \textsc{History} are evaluated separately: a question can have annotated evidence in \textsc{Active} while its answer still changes when historical context is included.

\subsection{Additional Write-Path Controls}
\label{app:write_controls}

Table~\ref{tab:write_controls} provides additional MMB controls omitted from the main paper. The semantic-only $B=24$ condition changes both channel availability and candidate budget and is therefore not the controlled comparison used for our graph-channel claim. We also report the narrower deterministic rule that treats only measurements and doses as supersedable. This rule archives only 6.8\% of memories and achieves 16.4\% supersession-pair recall, demonstrating that a manually specified set of supersedable dimensions fails to cover much of the benchmark's evolving state.

\begin{table}[t]
\centering
\small
\setlength{\tabcolsep}{4.5pt}
\renewcommand{\arraystretch}{1.05}
\caption{Additional MMB write-path controls. Impact ceiling measures coverage of the learned impact-retrieval candidate set and is therefore not applicable to deterministic supersession rules.}
\label{tab:write_controls}
\begin{tabular}{lrrrr}
\toprule
Variant & Archived & Pair recall & False arch. & Impact ceiling \\
\midrule
Full, semantic+graph, $B=48$ & 34.8 & 53.9 & 7.6 & 71.4 \\
Semantic only, $B=48$ & 33.0 & 51.6 & 12.3 & 66.7 \\
Semantic only, $B=24$ & 25.6 & 42.1 & 17.6 & 52.6 \\
Rule: measurement/dose & 6.8 & 16.4 & 8.6 & N/A \\
Rule: all non-episodic concepts & 20.6 & 28.6 & 17.6 & N/A \\
\bottomrule
\end{tabular}
\end{table}

\section{Deletion-Gate Calibration and Diagnostics}
\label{app:deletion_gate}

\subsection{Endurance Labels}
\label{app:deletion_gold}

Deletion is defined over three endurance classes. \emph{Safe-to-forget} memories are resolved and no longer expected to affect future reasoning; \emph{never-forget} memories remain clinically or temporally relevant despite being historical; and \emph{uncertain} memories cannot be safely classified from their content alone. The deletion rule treats both never-forget and uncertain memories as protected.

We construct labels at the concept-type level rather than the individual-memory level to prevent near-duplicate atoms of the same type from leaking across training and evaluation splits. The resulting MedMemoryBench label set contains 7,073 concept types: 2,849 safe to forget, 3,715 uncertain, and 509 never forget. Labels are produced by a blinded LLM judge from the concept and up to three example memories. All deletion-scorer training, calibration, and evaluation reported below use these LLM-derived endurance labels.

\subsection{Deletion Scorer and Operating Point}
\label{app:deletion_scorer}

The deletion scorer uses \texttt{bge-small-en-v1.5} with a binary classification head over the rendered concept and memory content. Safe-to-forget examples receive the positive label, while never-forget and uncertain examples are treated as protected. Splits are performed by concept type, yielding 1,595 training, 679 development, and 1,119 calibration types.

We select the deletion threshold using the calibrated safety constraint rather than by maximizing deletion recall. The resulting operating point is $\tau=0.988$. Applied to the deployment store, this threshold removes 2.3\% of stored memories, with 16.1\% safe-to-forget recall, 92.9\% safe precision, and 0.3\% never-forget loss.

\subsection{Conformal Risk Control}
\label{app:deletion_certificate}

For a protected concept type $t$, let $L_t(\tau)=\frac{1}{|M_t|}\sum_{m\in M_t}\mathbb{I}[s(m)\geq\tau]$, where $s(m)$ is the deletion score and $M_t$ is the set of memories belonging to type $t$. For safe-to-forget types, the loss is defined as zero. We select the smallest threshold satisfying $\frac{\sum_{t=1}^{n}L_t(\tau)+1}{n+1}\leq\alpha$, with $\alpha=0.01$ on the sealed calibration set.

On the sealed calibration set, the strict criterion selects $\tau=0.988$. At this threshold, 445 of 7,887 calibration memories are deleted, including 4 of 423 never-forget memories and 41 uncertain memories, while 11.1\% of safe-to-forget memories are cleared. We then apply this fixed threshold to 3,483 archived concept types from the deployment store; relative to the LLM-derived endurance labels, this operating point removes 2.3\% of the store with 16.1\% safe recall, 92.9\% safe precision, and 0.3\% never-forget loss (Table~\ref{tab:deletion_loss_definition}).

The resulting guarantee should be interpreted narrowly. Under exchangeability with the calibration distribution, the criterion controls the marginal expected deletion loss over concept types at \(\alpha=0.01\). Because safe-to-forget types contribute zero loss, this does not constitute a 1\% class-conditional bound on deletion loss among protected types. This guarantee is defined with respect to the LLM-derived endurance labeling scheme over unseen concept types; it does not bound deletion precision, per-memory clinical risk, or medical safety.

\subsection{Why the Strict Loss Is Necessary}
\label{app:deletion_loss_ablation}

We additionally calibrate a criterion that protects only the never-forget class. Because never-forget types constitute a small minority of all calibration types, averaging this loss over the complete calibration set substantially dilutes their contribution. The resulting threshold, $\tau=0.106$, deletes 49.6\% of calibration memories and clears 72.5\% of safe-to-forget memories, but also deletes 23.4\% of never-forget memories. On the deployment sweep, the same operating point produces an 8.1\% never-forget loss.

\begin{table}[t]
\centering
\small
\setlength{\tabcolsep}{4.2pt}
\renewcommand{\arraystretch}{1.05}
\caption{Effect of the calibrated loss definition on deployment-store deletion. The never-forget-only criterion is shown as a diagnostic and is not used by the system.}
\label{tab:deletion_loss_definition}
\begin{tabular}{lrrrr}
\toprule
Criterion
& Store removed
& Safe recall
& Safe precision
& NF loss \\
\midrule
Strict protected loss
& 2.3\%
& 16.1\%
& 92.9\%
& \textbf{0.3\%} \\
Never-forget only
& 14.7\%
& 76.3\%
& 69.4\%
& 8.1\% \\
\bottomrule
\end{tabular}
\end{table}

This comparison shows why the protected class used for calibration must match the operational deletion rule. In particular, a marginal risk bound over all concept types can provide a formally valid but operationally weak guarantee when the safety-critical class is rare.

\subsection{Deletion Candidate and Safety Ablations}
\label{app:deletion_safety_ablations}
\label{app:deletion_proposal}

\paragraph{Temporal status and memory endurance.}
We test a deterministic heuristic that removes archived memories whose extractor-assigned status is neither \texttt{ongoing} nor \texttt{historical}, while retaining diagnoses. It removes 26.4\% of the store and captures 83.4\% of memories labeled safe to forget, but its safe precision is only 42.3\%. It also removes 264 of 617 never-forget memories. These results show that temporal status alone does not establish whether a memory can be removed under the endurance-labeling scheme.

\paragraph{Deletion candidate sources.}
We examine explicit updater \textsc{Delete} proposals and a sweep over archived memories as candidate sources. The updater-proposal diagnostic covers 12 personas and yields 26 deletions under the strict gate. The archived-memory sweep yields 562 deletions. Table~\ref{tab:deletion_candidate_ablation} reports the candidate counts and deletion metrics for these runs. The 12-persona proposal diagnostic has a different cohort scope from the downstream QA evaluation on personas 1--5.

\begin{table}[t]
\centering
\small
\setlength{\tabcolsep}{4.2pt}
\renewcommand{\arraystretch}{1.07}
\caption{Reported deletion diagnostics by candidate source. Both runs use the same deletion gate; the updater-proposal diagnostic covers 12 personas.}
\label{tab:deletion_candidate_ablation}
\begin{tabular}{lrrrr}
\toprule
Candidate source
& Candidates
& Safe recall
& Safe precision
& Store removed \\
\midrule
Updater \textsc{Delete} proposals
& 535
& 10.2\%
& 84.6\%
& 0.05\% \\
All archived memories
& 8,418
& 16.1\%
& 92.9\%
& 2.3\% \\
\bottomrule
\end{tabular}
\end{table}

Restricting deletion candidates to explicit updater proposals prevents the gate from considering other archived memories. The framework therefore evaluates archived candidates independently of whether the updater proposed deletion, while keeping temporal state updates separate from endurance-based removal.

\subsection{Qualitative Examples and Threshold Limitations}
\label{app:deletion_failure}
\label{app:deletion_examples}

Examples removed by the strict gate include transient behavioral or short-lived state descriptions, such as a temporary humidifier setting, short-term throat dryness, planned soup-consumption timing, and a next-day wake-up report. Retained examples include ongoing medication use, persistent exercise intolerance, and longitudinal monitoring states.

Protected examples near the deletion threshold include fasting C-peptide measurements, six-minute walk distance, CPAP pressure settings, and prior reports of shortness of breath. Their potential value as longitudinal baselines distinguishes them from transient observations with similar wording. Conversely, some high-scoring memories remain below the deletion threshold despite appearing ephemeral, including short-term symptom reports and pillow-position recommendations. These examples suggest that additional context about longitudinal relevance may help distinguish transient observations from protected memories.

\subsection{End-to-End Effect of Deletion}
\label{app:deletion_e2e}

Table~\ref{tab:deletion_e2e} reports the complete paired downstream evaluation after applying the strict deletion gate. Both stores are served and answered in the same reader session to avoid interpreting cross-session model variation as an effect of deletion.

\begin{table}[t]
\centering
\small
\setlength{\tabcolsep}{3.8pt}
\renewcommand{\arraystretch}{1.05}
\caption{Paired MedMemoryBench evaluation before and after strict deletion.}
\label{tab:deletion_e2e}
\begin{tabular}{lrrrrrrrr}
\toprule
Store
& Macro
& Pooled
& EEM
& IG
& MCD
& MQ
& SUA
& TLA \\
\midrule
Untouched
& 36.82
& 39.11
& 63.0
& 22.0
& 8.5
& 41.4
& 44.0
& 42.0 \\
After deletion
& 37.78
& 40.32
& 63.0
& 24.0
& 4.3
& 41.4
& 48.0
& 46.0 \\
\bottomrule
\end{tabular}
\end{table}

Across all 496 questions, 20 answers improve and 14 regress after deletion, yielding exact McNemar $p=0.39$. A total of 287 answers are byte-identical across the two conditions. The aggregate score therefore does not indicate a clear loss in downstream
answer quality, and we do not interpret the nominal increase after deletion as
an improvement.

\section{Serving Efficiency and Computational Cost}
\label{app:serving_efficiency}

This section characterizes the computational cost of selective HISTORY
serving. We report serving-arm prompt length and TTFT, reader FLOPs,
KV-cache capacity and request-level occupancy, and the write-time cost of
constructing the structured memory.

\subsection{Measurement Protocol and Scope}
\label{app:serving_protocol}
\label{app:serving_scope}

Reader latency is measured using \texttt{Qwen/Qwen3-4B-Instruct-2507} in bfloat16, served with vLLM 0.22.0~\citep{kwon2023vllm} on one NVIDIA A100-SXM4-40GB GPU. Requests are executed sequentially at batch size one with temperature zero and prefix caching disabled. Disabling prefix caching prevents interleaved serving conditions from reusing previously computed prefixes. The results characterize this serving configuration; cache-aware workloads and concurrent-serving throughput are not evaluated.

We replay the reader inputs for joint-store and \textsc{Active}-only serving. MedMemoryBench, MIMIC-QA, and MedLoCoMo use saved request packets. ClinicalBench prompts are reconstructed from stored serving inputs using the original renderer. These measurements exclude memory construction, retrieval, and gate evaluation. Prompt-token counts are obtained from the server's usage accounting.

TTFT is measured from request submission to receipt of the first generated content token. The primary experiment uses a one-token completion budget and three paired, interleaved sweeps. We take the median measurement for each question and serving condition before aggregating across questions. The evaluation includes 496 MedMemoryBench, 400 MIMIC-QA, 400 ClinicalBench, and 320 MedLoCoMo questions, yielding 9,696 measured requests across the two serving conditions and three sweeps.

We compare contexts served to the same reader under the same execution stack. We do not report controlled wall-clock comparisons against other memory systems because the available baseline runtime measurements use different hosts and serving configurations. KV-cache allocation and single-request occupancy are measured separately; prompt-only capacity estimates are not measurements of concurrent throughput.

\subsection{Serving-Arm Latency}
\label{app:serving_arm_latency}

Table~\ref{tab:serving_arm_latency} first characterizes the two serving arms independently of any gating policy. \textsc{Both} always serves retrieved memories from both \textsc{Active} and \textsc{History}, whereas \textsc{Active}-only omits historical memories. Because both arms are replayed for every question, these measurements describe the computational consequence of reducing the served context without depending on which gate selects that arm.

\begin{table}[t]
\centering
\small
\setlength{\tabcolsep}{5.0pt}
\renewcommand{\arraystretch}{1.06}
\caption{Prompt length and TTFT. Values are means over the one-token prefill experiment. }
\label{tab:serving_arm_latency}
\begin{tabular}{lrrrr}
\toprule
Dataset & \multicolumn{2}{c}{\textsc{Both}} & \multicolumn{2}{c}{\textsc{Active}-only} \\
\cmidrule(lr){2-3}
\cmidrule(lr){4-5}
& Tokens & TTFT & Tokens & TTFT \\
\midrule
MedMemoryBench & 4,482 & 228.7 ms & 1,876 & 97.3 ms \\
MIMIC-QA & 2,484 & 125.3 ms & 1,044 & 59.9 ms \\
ClinicalBench & 1,052 & 55.9 ms & 653 & 41.9 ms \\
MedLoCoMo & 1,680 & 90.9 ms & 1,478 & 84.1 ms \\
\bottomrule
\end{tabular}
\end{table}

The difference between the two arms is largest when \textsc{History} contributes a substantial fraction of the prompt. For example, the average MedMemoryBench prompt decreases from 4,482 to 1,876 tokens, while mean TTFT decreases from 228.7 to 97.3~ms. On MedLoCoMo, the corresponding context difference is much smaller, from 1,680 to 1,478 tokens, and consequently the latency difference between the two serving actions is also smaller.

\subsection{Policy-Level TTFT Savings}
\label{app:policy_latency}

The latency benefit of selective HISTORY serving depends on both how often
the gate selects the smaller ACTIVE-only packet and how much HISTORY
contributes to the prompt for those queries. The serving-arm measurements
show that omitting HISTORY can substantially reduce prefill TTFT when it
accounts for a large fraction of the prompt, while the benefit is smaller
when the two serving arms have similar context lengths.

\subsection{Decode and End-to-End Behavior}
\label{app:decode_latency}

The store gate directly changes the prefilled context but does not change the decoding algorithm. We verify this distinction on a deterministic 100-question subset from each dataset using the original generation budget, with both serving actions evaluated in three sweeps.

\begin{table}[t]
\centering
\small
\setlength{\tabcolsep}{4.5pt}
\renewcommand{\arraystretch}{1.06}
\caption{Decode behavior on the deterministic 100-question generation subset.}
\label{tab:decode_latency}
\begin{tabular}{lrrrr}
\toprule
Dataset
& \multicolumn{2}{c}{Completion tokens}
& \multicolumn{2}{c}{Decode time} \\
\cmidrule(lr){2-3}
\cmidrule(lr){4-5}
& \textsc{Both}
& \textsc{Active}-only
& \textsc{Both}
& \textsc{Active}-only \\
\midrule
MedMemoryBench & 102.8 & 99.5  & 0.877 s & 0.821 s \\
MIMIC-QA       & 53.2  & 52.8  & 0.432 s & 0.419 s \\
ClinicalBench  & 189.5 & 146.5 & 1.550 s & 1.187 s \\
MedLoCoMo      & 8.8   & 8.7   & 0.056 s & 0.054 s \\
\bottomrule
\end{tabular}
\end{table}

Decode throughput remains approximately 120--150 tokens/s under both serving actions, confirming that the direct computational benefit of store selection is primarily a prefill effect. ClinicalBench is unusual because the smaller context also changes generation length, reducing the mean completion from 189.5 to 146.5 tokens. We therefore do not interpret its shorter decode time as a pure serving-efficiency gain and use TTFT rather than end-to-end latency as the primary latency metric.

\subsection{Reader FLOP Accounting}
\label{app:reader_flops}

We additionally report FLOPs as a hardware-independent measure of reader computation, counting one multiply-accumulate as two FLOPs. A linear layer applied to a sequence of length $T$, with input dimension $n$ and output dimension $m$, therefore contributes $\mathrm{FLOPs}_{\mathrm{linear}} = 2Tnm$.
Causal attention is counted over the lower-triangular attention matrix. Decode computation assumes a KV cache and sums the attended context over generated positions. The language-model head is counted at the final position during prefill and at every generated position during decoding. Embedding lookup, normalization, RoPE, softmax, and other elementwise operations are excluded. The reader contains 36 layers with hidden dimension 2,560, FFN dimension 9,728, 32 query heads, 8 KV heads, and head dimension 128, for approximately 4.022 billion parameters.

\begin{table}[t]
\centering
\small
\setlength{\tabcolsep}{5.2pt}
\renewcommand{\arraystretch}{1.06}
\caption{Prefill FLOPs of the two serving arms. These values characterize the computational cost of presenting each context to the reader and are independent of a gate's commit frequency.}
\label{tab:serving_arm_flops}
\begin{tabular}{lrrr}
\toprule
Dataset & \textsc{Both} & \textsc{Active}-only & Reduction \\
\midrule
MedMemoryBench & 38.5 TFLOP & 14.7 TFLOP & 61.9\% \\
MIMIC-QA & 19.9 TFLOP & 7.9 TFLOP & 60.2\% \\
ClinicalBench & 8.0 TFLOP & 4.9 TFLOP & 38.9\% \\
MedLoCoMo & 13.0 TFLOP & 11.4 TFLOP & 12.7\% \\
\bottomrule
\end{tabular}
\end{table}

The FLOP reduction can exceed the token reduction because shortening the prompt also reduces the attention computation. For example, the MedMemoryBench \textsc{Active}-only arm contains 58.2\% fewer prompt tokens than \textsc{Both}, while its estimated prefill computation is 61.9\% lower. Unlike the wall-clock measurements, which depend on the A100 serving stack, these FLOP and token counts do not depend on a particular accelerator.

\subsection{KV-Cache Capacity and Occupancy}
\label{app:kv_cache}

With the serving configuration above, vLLM reports a KV-cache capacity of 186,288 tokens and 25.58~GiB of available KV memory, implying approximately 143.98~KiB per cached token. This agrees with the analytic bfloat16 requirement, $2 \times 36 \times 8 \times 128 \times 2 = 147{,}456$ bytes $= 144.00$~KiB/token.
where the factors represent keys and values, layers, KV heads, head dimension, and bytes per value. Table~\ref{tab:kv_cache} reports the resulting request-level footprints and prompt-only capacity estimates.


\begin{table*}[t]
\centering
\small
\setlength{\tabcolsep}{5pt}
\renewcommand{\arraystretch}{1.06}
\caption{KV-cache footprint and prompt-only capacity under the two serving actions. Capacity is computed from the measured 186,288-token cache and ignores generated tokens and 16-token block rounding; it is therefore an upper bound on simultaneous requests, not a throughput measurement.}
\label{tab:kv_cache}
\begin{tabular}{lrrrr}
\toprule
Dataset
& \textsc{Both}
& \textsc{Active}-only
& KV reduction
& Prompt-only capacity \\
\midrule
MedMemoryBench & 630 MiB & 264 MiB & 58\% & $41.6 \rightarrow 99.3$ \\
MIMIC-QA       & 349 MiB & 147 MiB & 58\% & $75.0 \rightarrow 178.4$ \\
ClinicalBench  & 148 MiB &  92 MiB & 38\% & $177.1 \rightarrow 285.3$ \\
MedLoCoMo      & 236 MiB & 208 MiB & 12\% & $110.9 \rightarrow 126.0$ \\
\bottomrule
\end{tabular}
\end{table*}

We validate request-level occupancy using 40 requests, comprising five questions per dataset and serving condition, with one request in flight and a forced 128-token completion. Sampling \texttt{vllm:kv\_cache\_usage\_perc} every 20~ms gives a measured-to-expected occupancy ratio with median 1.0019 and range 0.9976--1.0100. Mean excess occupancy ranges from 1.5 to 7.6 tokens across conditions, consistent with 16-token block allocation. The prompt-only capacity estimates exclude generated tokens and block rounding and should therefore be interpreted as memory-capacity estimates rather than measured throughput.

\subsection{Write-Time Construction Cost}
\label{app:write-cost}

\begin{table}[H]
\centering
\small
\setlength{\tabcolsep}{3pt}
\renewcommand{\arraystretch}{1.05}

\begin{minipage}[t]{0.49\linewidth}
\vspace{0pt}
\centering
\textit{(a) Wall-clock construction}\\[3pt]

\resizebox{\linewidth}{!}{
\begin{tabular}{llrr}
\toprule
Dataset & Record & Source tok. & Time \\
\midrule
MedMemoryBench
& shortest & 647.6K & 444 s \\
& median   & 668.1K & 538 s \\
& longest  & 818.1K & 595 s \\
\midrule
ClinicalBench
& shortest & 3.4K  & 54 s \\
& median   & 15.1K & 402 s \\
& longest  & 73.1K & 2,128 s \\
\bottomrule
\end{tabular}
}
\end{minipage}
\hfill
\begin{minipage}[t]{0.49\linewidth}
\vspace{0pt}
\centering
\textit{(b) MedMemoryBench model compute}\\[3pt]

\resizebox{\linewidth}{!}{
\begin{tabular}{lrr}
\toprule
Stage & Calls & Compute \\
\midrule
Atomic extraction & 2,646 & 37.75 PFLOP \\
Updater agent & 500 & 4.21 PFLOP \\
Relation linking & 18,900 & 50.04 PFLOP \\
\midrule
Total & 22,046 & 92.00 PFLOP \\
\bottomrule
\end{tabular}
}
\end{minipage}
\caption{
Write-time construction cost. Panel (a) reports end-to-end wall-clock latency on representative records using Qwen3-1.7B on a single otherwise idle GPU. Panel (b) reports model compute for the published Qwen3-1.7B MedMemoryBench stores over the five evaluation personas, including atomic extraction, updating, and relation linking.
}
\label{tab:write_cost_summary}
\end{table}

We characterize write-time cost in two complementary ways:
end-to-end wall-clock construction latency on representative records
and model-call, token, and FLOP accounting from the recorded
construction workload. The former measures realized pipeline latency,
whereas the latter characterizes model computation independently of
concurrent execution.

Construction cost varies substantially across benchmarks and is not
proportional to raw record length. MedMemoryBench processes records
exceeding 600K source tokens at 0.69--0.81 seconds per 1K tokens,
whereas ClinicalBench requires 16.0--29.1 seconds per 1K tokens.
The difference reflects the number and structure of write-time model
operations induced by the source record rather than source-token
volume alone.

Memory construction incurs a one-time write cost. For the published
MedMemoryBench evaluation stores, atomic extraction, state updating, and
relation linking require 37.75, 4.21, and 50.04 PFLOP, respectively,
for 92.00 PFLOP in total. These costs are incurred during memory
construction rather than at every downstream query; their practical
amortization therefore depends on the update frequency and the number of
subsequent queries served from the maintained memory.


\end{document}